\documentclass[journal]{IEEEtran}
\usepackage{graphicx}
\usepackage{epsfig}
\usepackage{latexsym}
\usepackage{amsfonts}
\usepackage{here}
\usepackage{rawfonts}
\usepackage[latin1]{inputenc}
\usepackage[T1]{fontenc}
\usepackage{calc}
\usepackage{url}
\usepackage{enumerate}
\usepackage{color}
\usepackage[tbtags]{amsmath}
\usepackage{amssymb}
\usepackage{upref}
\usepackage{epic,eepic}
\usepackage{times}
\usepackage{dsfont}
\usepackage{comment}
\usepackage{cite}

\usepackage{graphicx}
\usepackage[font={small}]{caption}
\usepackage[font={small}]{subcaption}

\usepackage{stfloats}
\usepackage{mathtools}
\usepackage{amsmath}

\usepackage{algorithm}
\usepackage{algpseudocode}

\newtheorem{theorem}{\bf{Theorem}}
\newtheorem{definition}{\bf{Definition}}
\newtheorem{corollary}{\bf{Corollary}}
\newtheorem{lemma}{\bf{Lemma}}
\newtheorem{remark}{Remark}

\ifCLASSINFOpdf
\else
\fi
\begin{document}

	\title{A Hierarchical Federated Learning Approach for the Internet of Things}
	
	\author{{Seyed Mohammad Azimi-Abarghouyi and Viktoria Fodor}

		\thanks{The authors are with the School of Electrical Engineering and Computer Science and Digital Futures, KTH Royal Institute of Technology, Stockholm, Sweden (Emails: $\bigl\{$seyaa,vjfodor$\bigr\}$@kth.se).}
	}

	\maketitle
	
	\begin{abstract}
		This paper presents a novel federated learning solution, QHetFed, suitable for large-scale Internet of Things deployments, addressing the challenges of large geographic span, communication resource limitation, and data heterogeneity. QHetFed is based on hierarchical federated learning over multiple device sets, where the learning process and learning parameters take the necessary data quantization and the data heterogeneity into consideration to achieve high accuracy and fast convergence.  Unlike conventional hierarchical federated learning algorithms, the proposed approach combines gradient aggregation in intra-set iterations with model aggregation in inter-set iterations. We offer a comprehensive analytical framework to evaluate its optimality gap and convergence rate, and give a closed form expression for the optimal learning parameters under a deadline, that accounts for communication and computation times. Our findings reveal that QHetFed consistently achieves high learning accuracy and significantly outperforms other hierarchical algorithms, particularly in scenarios with heterogeneous data distributions.
		
	\end{abstract}
	

	\maketitle
	
	\section{Introduction}
	
	\IEEEPARstart{F}{ederated} learning (FL) in Internet of Things (IoT) deployments makes it possible to learn from highly distributed data, without costly data transmission and under privacy constraints \cite{mcmahan, viktoria}. It is also an efficient approach to speed up the learning process, since learning is performed simultaneously at several devices \cite{gupta}. However, efficient FL in the IoT scenario is challenged by the large geographic span of the deployment, and the typically limited networking resources of the devices. In addition, sets of devices may belong to different authorities, and the data they possess can be highly heterogeneous, as it originates from diverse environments.
	
	The key learning approach in this scenario is hierarchical FL, where sets of devices
	perform  one or more local learning rounds, and then exchange and aggregate  model parameters or gradients via local edge servers. The edge servers then collaborate again, most typically by aggregating model parameters at a cloud server.
	
	The use of this hierarchical structure has been proposed for and can be beneficial in several scenarios. The most obvious reason to implement hierarchical structures is the large geographic span of the involved devices. 
	In wireless networks, a hierarchical structure can improve the quality of the transmissions over the wireless channels \cite{wen}, 
	and localizing the part of the learning saves communication resources and time \cite{gupta,tony}.
	Clustering devices could be useful also to deal with device or network heterogeneity \cite{lin,wu}, 
	keep data traffic localized within an administrative unit or social groups \cite{zhou}, 
	or simply adjust to the topology of the interconnections in mobile networks, over the internet, or in computing infrastructures \cite{zhang, gupta}. Additionally, hierarchical structures are crucial for big data, enabling efficient and scalable processing of large datasets while reducing communication burdens \cite{bigdata}.
	
	In this paper, we propose a hierarchical FL solution that specifically addresses the challenges of FL in IoT systems, by accounting for the effects of potentially severe data quantization and the data heterogeneity among devices.
	
	\subsection{State of the Art}
	The main challenges to achieve efficient and effective FL in a single cell \cite{smith} are non-i.i.d. or heterogeneous data distribution \cite{bennis}, known as data or statistical heterogeneity, heterogeneous devices and networks \cite{tran}, known as systems heterogeneity, and the efficient use of network resources \cite{amiri_basic_ota}. Hierarchical FL comes with additional challenges. The placement of the aggregators and the optimal formation of the device sets are addressed in \cite{bennis,tony,place}, and the sharing of resources among parallel FL sessions is discussed in \cite{lim, energy_resource, cost, context,scheduling}. A unified clustering method is suggested in \cite{feduc}. The limited transmission capacity is considered in several works, for example \cite{wen,gunduz_ota}, while \cite{castiglia,zhang} focus on the scenario with limited connectivity among the edge servers and \cite{lin} addresses the scenario of limited edge-to-cloud network resources. Learning based resource allocation under dynamically changing computing and transmission resources are considered in \cite{su}. The effect of the distortion of the transmitted model and gradient parameters due to wireless inter-cell interference is considered in \cite{ourjournal, ourconf}, while \cite{letaief} evaluates the consequences of quantization on the learning convergence. These works show that the distortion of the parameters leads non-diminishing learning loss. As another bandwidth-limited approach, \cite{prun1,prun2} propose using pruning to reduce the scale of the neural network.
	
	Local gradient descent, aggregation at the edge, and aggregation at the cloud can be organized in various ways. As of now, no general results are available that dictate a specific learning structure. In \cite{fedsgd-is-better}, various combinations of local gradient descent, gradient parameter aggregation, and model parameter aggregation at the edge are compared within a single-cell FL. It is shown that initially employing multiple-step gradient descent with model aggregation at the edge, followed by iterations of single-step gradient descent with gradient aggregation, yields best performance among the considered combinations.
	In hierarchical FL, model parameters are aggregated at both the edge servers and the cloud server for example in \cite{castiglia,wang,letaief}, while gradient aggregation is applied on both levels in \cite{tony,wen}. The mix of gradient and model parameter aggregation is proposed in \cite{ourjournal, ourconf}, where gradient aggregation is performed at the intra-set iterations and model aggregation at the inter-set iterations. The scheme is deployed for interference-limited scenarios. It is demonstrated that the approach leads to convergent learning, when high interference leads to significant uplink and downlink transmission errors, and conventional hierarchical FL \cite{letaief} with multiple-step gradient descent and model aggregation at both edge server and cloud server becomes unstable.
	
	While most of the research on hierarchical FL \cite{lin, wu, zhou, bennis, place, lim, energy_resource, cost, context, scheduling, feduc, gunduz_ota, su, letaief, prun1, prun2} has focused solely on model parameter aggregation, its ability to operate effectively under data heterogeneity is limited, as discussed in \cite{fedsgd-is-better}. The challenge of data heterogeneity is markedly more pronounced in hierarchical systems, where the number of devices involved in the learning process can be much higher than in single-cell FL, and devices may be distributed across different geographic regions or belong to specific communities. This highlights the need for a new aggregation approach in hierarchical FL.
	
	\vspace{0pt}
	\subsection{Contributions} 
This work extends the state of art, by introducing a novel hierarchical FL framework that tackles the challenges of data heterogeneity inherent in large-scale IoT deployments, and noisy data transmission as the consequence of quantization \cite{letaief, alish, ramtin, qjsac}. Our main contributions are outlined below:
	
	\textit{Learning Approach:} We propose a new iterative learning
	method called {\fontfamily{lmtt}\selectfont {QHetFed}}, that combines intra-set gradient and
	inter-set model parameter aggregations, together with
	multiple-step gradient descent at the end of each inter-set
	iteration to expedite the learning procedure.
	Based on the results of \cite{ourjournal, ourconf}, we expect this approach to exhibit strong
	resilience to non-i.i.d. data and quantization noise.
	
	\textit{Heterogeneity-Aware Convergence Analysis:} We derive the optimality gap parameterized by quantization factors and a data
	heterogeneity metric.  Notably, our analysis
	shows that the optimality gap grows independently with both
	data heterogeneity and the variance of quantization error. We
	extend the analysis of conventional hierarchical FL in
	\cite{letaief} to cover heterogeneous data, and discuss the potential of the two schemes.
	We provide practical remarks to aid system and learning algorithm design. 
	
	\textit{System Optimization:} We derive the convergence rate of our method and use this finding to formulate an optimization problem to determine the optimal numbers of intra-set iterations and gradient descent steps under runtime deadline. The optimal values take the communication and computation times as well as the variance of quantization error into account, and are expressed in closed-form.
	
	\textit{Insights:} The analytical and experimental results demonstrate
	that {\fontfamily{lmtt}\selectfont {QHetFed}} is superior over its conventional
	hierarchical counterpart under heterogeneous data
	distributions and limited quantization, while has slightly slower
	convergence under  homogeneous data.
	Our analysis also reveals that the parameters of the learning
	algorithm need to be set by taking the quantization levels as well as the
	maximum and minimum number of devices per set into account.

	\section{Proposed Hierarchical Scheme}
	In situations where gradient and model parameters are affected by quantization noise or data heterogeneity, 
	the consequent errors tend to amplify through successive local steps. This is because each step involves computations on imprecise or altered parameters. Specifically, near the optimum, some device gradients might diverge from the optimum, as the local models approach the local optimal solutions instead of the global one. 
	Similar phenomena, highlighting significant performance declines in {\fontfamily{lmtt}\selectfont {{FedAvg}}} \cite{mcmahan} under noisy conditions or with non-i.i.d. data, are documented in \cite{hashemi,fedsgd-is-better}. Conversely, {\fontfamily{lmtt}\selectfont {QHetFed}}, the learning algorithm proposed in this work, implements a single local step in intra-set iterations, where the gradient is derived from aggregated data rather than local computations, potentially mitigating the impact of noise or deviations. We draw inspiration from \cite{huang_sg, yang3}, which showcases the resilience of gradient aggregation against interference, and from \cite{fedsgd-is-better} that demonstrates its effectiveness with non-i.i.d. data. {\fontfamily{lmtt}\selectfont {QHetFed}} strategically performs multiple-step local training only at the end of each inter-set iteration, just prior to a robust cloud aggregation that encompasses all participating devices across all sets.
	\subsection{Learning Algorithm} Assume that there are one cloud server, $C$ edge servers with disjoint device sets $\left\{{\cal C}^l\right\}_{l=1}^{C}$, each set ${\cal C}^l$ including $N_l$ devices with distributed datasets $\left\{{\cal D}_n^l\right\}_{n=1}^{N_l}$, as shown in Fig. 1.\footnote{In line with \cite{lin, castiglia, letaief, ourjournal, ourconf}, network optimization problems such as device selection, resource allocation, and clustering are beyond the scope of this work. Our focus is on proposing a new FL algorithm for a predefined hierarchical network architecture and examining the effects of quantization and data heterogeneity on performance. These issues can be explored in future works once the algorithm and its characteristics for any network architecture are established in this study. Additionally, please note that due to geographic constraints, there may be only a single possible network architecture.} The distributed datasets for each set or each device can generally be statistically different, as the devices may observe different environments and belong to different communities.
	
	The learning model is parametrized by the parameter vector $\mathbf{w} \in \mathbb{R}^d$, where $d$ denotes the learning model size. Then, the local loss function of the model parameter vector $\mathbf{w}$ over ${\cal D}_n^l$ is
	\begin{align}
		\label{localloss}
		F_{n}^{l}(\mathbf{w}) =  \frac{1}{D_n^l}\sum_{\xi\in {\cal D}_{n}^{l}}^{}\ell(\mathbf{w},\xi),
	\end{align}
	where $D_n^l = |{\cal D}_{n}^l|$ is the dataset size and $\ell(\mathbf{w},\xi)$ is the sample-wise loss function that measures the
	prediction error of $\mathbf{w}$ on a sample $\xi$. Then, the global loss function on the distributed datasets $\cup_{l}\cup_{n} {\cal D}_{n}^{l}$ is computed as
	\begin{align}
		\label{lossfunction}
		F(\mathbf{w}) = \frac{1}{\sum_{l} \sum_{n} D_n^l}\sum_{l}^{} \sum_{n}^{} D_n^l F_{n}^{l}(\mathbf{w}).
	\end{align}
	Therefore, the goal of the learning process is to find a desired model parameter vector $\mathbf{w}$ that minimizes $F(\mathbf{w})$ as
	\begin{align}
		\label{globalopt}
		\mathbf{w}^* = \min_{\mathbf{w}} F(\mathbf{w}).
	\end{align}
	\begin{figure}[tb!]
		\hspace{-80pt}
		\includegraphics[width =5.8in]{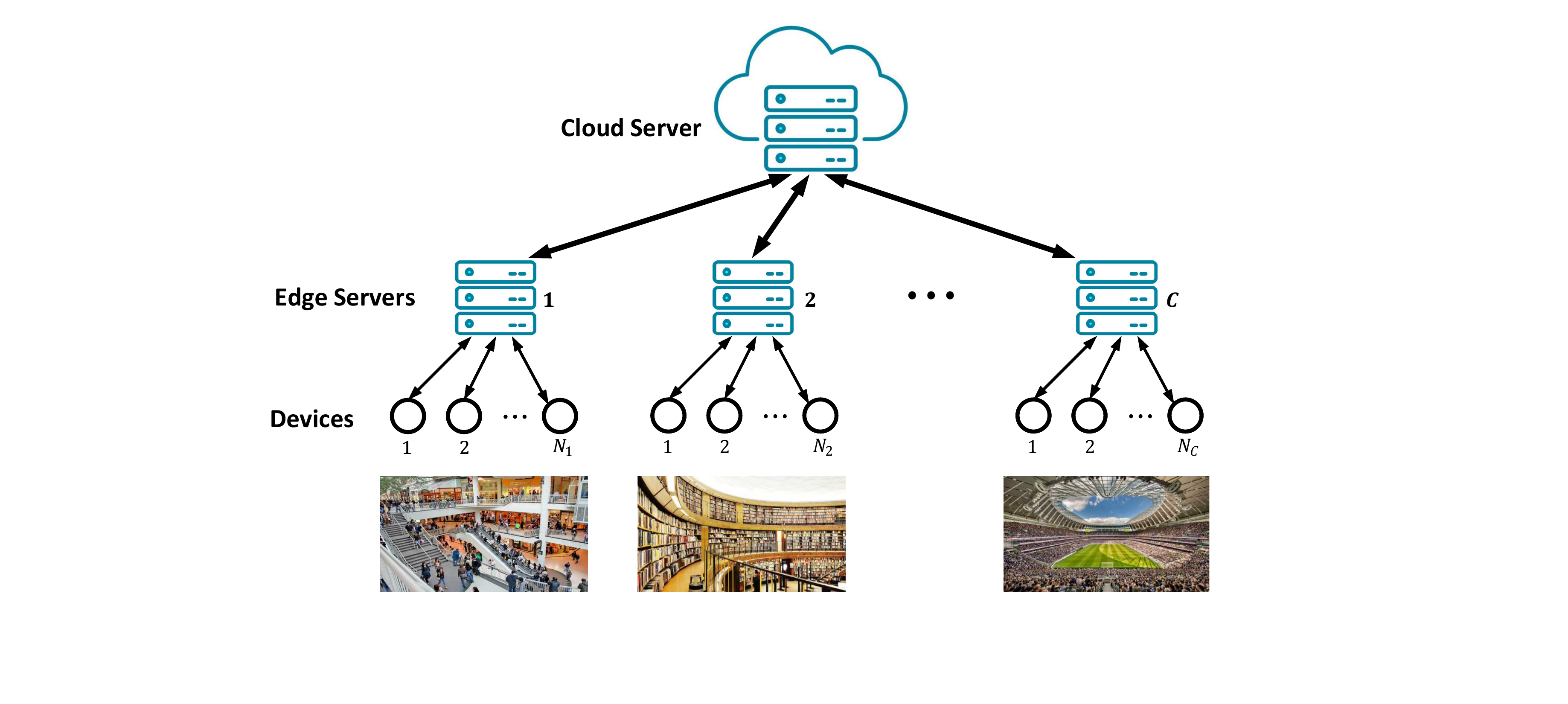}
		\vspace{-30pt}
		\caption{Hierarchical system. As an example of data heterogeneity, three different environments--- shopping mall, library, and stadium--- are illustrated for sets 1, 2, and $C$, respectively.}
		\vspace{0pt}
	\end{figure}
	We propose a new hierarchical algorithm called {\fontfamily{lmtt}\selectfont {QHetFed}} to tackle \eqref{globalopt}. Our approach involves two levels. Within $T$ global inter-set iterations, each iteration $t$ comprises $\tau$ intra-set iterations. During a specific intra-set iteration $i$, every device $n$ in a set $l$ computes the local gradient of the loss function in \eqref{localloss} from its local dataset, identified by the indices \(\left\{i,t\right\}\), as
	\begin{align}
		\label{localgrad}
		\mathbf{g}_{n,i,t}^{l} = \nabla F_{n}^{l}(\mathbf{w}_n^l,\boldsymbol \xi_n^l),
	\end{align}
	where $\mathbf{w}_n^l$ is its parameter vector and ${\boldsymbol\xi}_{n}^l$ with the size $B$ is the local mini-batch
	chosen uniformly at random from ${\cal D}_{n}^{l}$. Then, devices apply a quantizer operator $Q_1(.)$ on their local gradients and upload the results to their edge servers for edge aggregation. For this, the server $l$ averages of the local gradients from its devices as
	\begin{align}
		\label{intragrad}
		\mathbf{g}_{i,t}^{l} = \frac{1}{N_l} \sum_{n \in {\cal C}^l}^{} Q_1(\mathbf{g}_{n,i,t}^{l}).
	\end{align}
	Following the broadcast of the edge aggregated gradients $\mathbf{g}_{i,t}^{l}, \forall l$ to their devices by the servers, each device $n$ within any set $l$ proceeds to update its local model by implementing a one-step gradient descent as
	\begin{align}
		\label{localcomp}
		\mathbf{w}_{n,i+1,t}^{l} = \mathbf{w}_{n,i,t}^{l} -\mu \mathbf{g}_{i,t}^{l},
	\end{align}
	where $\mu$ is the learning rate. Upon finishing $\tau$ intra-set iterations, every device then performs a $\gamma$-step gradient descent as
	\begin{align}
		\label{final_local00}
		\mathbf{w}_{n,\tau,0,t}^{l} = \mathbf{w}_{n,\tau,t}^{l},
	\end{align}
	\begin{align}
		\label{final_local01}
		\mathbf{w}_{n,\tau,j,t}^{l} = \mathbf{w}_{n,\tau,j-1,t}^{l} - &\mu\nabla F_{n}^{l}(\mathbf{w}_{n,\tau,j-1,t}^{l},\boldsymbol\xi_{n,\tau,j-1,t}^l), \nonumber\\& j = \left\{1,\cdots,\gamma\right\}.
	\end{align} 
	This local multiple-step update facilitates acceleration in the learning process. To start the global inter-set iteration, each device $n \in {\cal C}^l$ applies the quantizer operator $Q_1(.)$ on the
	difference between its updated
	model $\mathbf{w}_{n,\tau,\gamma,t}^l$ to $\mathbf{w}_{n,\tau,t}^{l}$ and uploads the result to its server. Consequently, each server $l$ calculates an intra-set model parameter vector using the following average
	\begin{align}
		\label{intergrad}
		\mathbf{w}_{t+1}^{l} = \mathbf{w}_{\tau,t}^{l}+\frac{1}{N_l}\sum_{n}^{} Q_1\left(\mathbf{w}_{n,\tau,\gamma,t}^{l}-\mathbf{w}_{n,\tau,t}^{l}\right).
	\end{align}
	where $\mathbf{w}_{\tau,t}^{l} = \mathbf{w}_{n,\tau,t}^{l}, \forall n \in {\cal C}^l$, this denotes what the edge server $l$ can track from \eqref{localcomp}. Then, each edge server  $l$ applies a quantizer operator $Q_2(.)$ on its model $\mathbf{w}_{t+1}^l$ subtracted from the current global model and forwards the result to the cloud server for cloud aggregation as
	\begin{align}
		\label{globalgrad}
		\mathbf{w}_{t+1} = \mathbf{w}_t+\frac{1}{N} \sum_{l}^{}  N_l Q_2\left(\mathbf{w}_{t+1}^{l} -\mathbf{w}_t \right).
	\end{align}
	where $N = \sum_{l=1}^{C} N_l$ is the total number of devices. Then, each device $n \in {\cal C}^l, \forall l$ updates $\mathbf{w}_{n,0,t+1}^l = \mathbf{w}_{t+1}$ for the next global iteration $t+1$. This global update synchronizes all the local training processes over different sets. This procedure is detailed in Algorithm 1.
	
	We describe the quantization functions $Q_i$, for $i= \left\{1,2\right\}$, with two parameters, the number of quantization levels $s_i$ and the variance of the quantization error $q_i$. We assume the following characteristics for these functions.
	
	\textbf{Assumption 1 (Unbiased Quantization):} The quantizer $Q$ is unbiased and its variance grows with the square of the $l_2$-norm of its argument, as
	\begin{align}
		\mathbb{E}\left\{Q(\mathbf{x})|\mathbf{x}\right\} = \mathbf{x},
	\end{align}
	\begin{align}
		\mathbb{E}\left\{\|Q(\mathbf{x})-\mathbf{x}\|^2|\mathbf{x}\right\} \leq q \|\mathbf{x}\|^2,
	\end{align}
	for any $\mathbf{x} \in \mathbb{R}^d$ and positive real constant $q$ as the variance of quantization error.
	
	\textbf{Example for Quantizer \cite{alish}.} For any variable $\mathbf{x}\in \mathbb{R}^d$, the
	quantizer $Q^s$: $\mathbb{R}^d \to \mathbb{R}^d$ is defined as below
	\begin{align}
		Q^s (\mathbf{x}) = \textrm{sign}\left\{\mathbf{x}\right\} \|\mathbf{x}\| \zeta(\mathbf{x},s),
	\end{align}
	where the $i$-th element of $\zeta(\mathbf{x},s)$, i.e., $\zeta_i(\mathbf{x}, s)$, is a random variable as
	\begin{align}
		\zeta_i(\mathbf{x},s) &= 
		\frac{l}{s}\ \textrm{with probability}\ 1-v\left({\frac{|x_i|}{\|\mathbf{x}\|}},s\right)\nonumber\\& \hspace{-4pt}\textrm{and} \
		\frac{l+1}{s}\ \textrm{with probability}\ v\left(\frac{|x_i|}{\|\mathbf{x}\|},s\right),
	\end{align}
	where $x_i$ is the $i$-th element of $\mathbf{x}$ and $v(a, s) = as-l$ for any $a\in[0, 1]$. In above, the tuning parameter $s$ corresponds to the number of quantization levels and $l \in [0, s)$ is an integer such that $\frac{|x_i|}{\|\mathbf{x}\|}\in [\frac{l}{s},\frac{l+1}{s}]$. As shown in \cite{alish}, the variance $q$ decreases with increasing $s$.
	
	\begin{algorithm}
		\small
		\caption{{\fontfamily{lmtt}\selectfont {QHetFed}} algorithm}
		\begin{algorithmic}
			\vspace{-1pt}
			\State Initialize the global model $\mathbf{w}_0$\
			\vspace{0pt}
			\State \textbf{for} inter-set iteration $t=1,...,T$ \textbf{do}
			\vspace{0pt}
			\State \hspace{10pt}Each device updates its model by $\mathbf{w}_{t}$ 
			\vspace{0pt}
			\State \hspace{10pt}\textbf{for} intra-set iteration $i=1,...,\tau$ \textbf{do}
			\vspace{0pt}
			\State \hspace{20pt}Each device obtains its local gradient from $\mathbf{g}_{n,i,t}^{l} = \nabla F_{n}^{l}(\mathbf{w}_{n,i,t}^l,\boldsymbol\xi_{n,i,t}^l)$
			\vspace{0pt}
			\State \hspace{20pt}Each edge server obtains its intra-set gradient from $\mathbf{g}_{i,t}^{l} = \frac{1}{N_l}\sum_{n \in {\cal C}^l}^{} Q_1(\mathbf{g}_{n,i,t}^{l})$
			\vspace{0pt}
			\State \hspace{20pt}Each device updates its local model as $\mathbf{w}_{n,i+1,t}^{l} = \mathbf{w}_{n,i,t}^{l} -\mu \mathbf{g}_{i,t}^{l}$
			\vspace{0pt}
			\State \hspace{20pt}\textbf{if} $i = \tau$ \textbf{do}
			\vspace{0pt}
			\State \hspace{30pt}Each device updates its local model as $\small{\mathbf{w}_{n,\tau,0,t}^{l} = \mathbf{w}_{n,\tau,t}^{l},\
				\mathbf{w}_{n,\tau,j,t}^{l} = \mathbf{w}_{n,\tau,j-1,t}^{l} - \mu}\times$\\\hspace{70pt}$\small{\nabla F_{n}^{l}(\mathbf{w}_{n,\tau,j-1,t}^{l},\boldsymbol\xi_{n,\tau,j-1,t}^l), \ j \leq\gamma}$
			\vspace{0pt}
			\State \hspace{10pt}Each edge server obtains its intra-set model from $\mathbf{w}_{t+1}^{l} = \mathbf{w}_{\tau,t}^{l}+\frac{1}{N_l}\sum_{n}^{} Q_1\left(\mathbf{w}_{n,\tau,\gamma,t}^{l}-\mathbf{w}_{n,\tau,t}^{l}\right)$
			\vspace{0pt}
			\State \hspace{10pt}Cloud server obtains global model from $\mathbf{w}_{t+1} = \mathbf{w}_t+\frac{1}{N} \sum_{l}^{}  N_l Q_2\left(\mathbf{w}_{t+1}^{l} -\mathbf{w}_t \right)$
		\end{algorithmic}
		
	\end{algorithm}	
	
	\subsection{Convergence Analysis}
	The theorem presented next provides the convergence performance of {\fontfamily{lmtt}\selectfont {QHetFed}} in terms of the optimality gap. This is contextualized within the framework of data heterogeneity and widely recognized assumptions prevalent in the literature, as detailed below.
	
	\begin{definition}
		The heterogeneity of the local data distributions ${\cal D}_n^l, \forall n, l$ is captured by a popular notion of data heterogeneity, $G^2$, defined as follows \cite{fedsgd-is-better}.
		\begin{align}
			G^2 = \max_{n, l} \sup_{\mathbf{w}} \|\nabla F(\mathbf{w}) - \nabla F_n^l(\mathbf{w})\|^2.
		\end{align}	
	\end{definition}
	
	\textbf{Assumption 2 (Lipschitz-Continuous Gradient):} The gradient of the loss function $F(\mathbf{w})$, as represented in \eqref{lossfunction}, exhibits Lipschitz continuity with a positive constant $L > 0$. This means that for every two model vectors $\mathbf{w}_1$ and $\mathbf{w}_2$, the following holds.
	\begin{align}
		&F(\mathbf{w}_2) \leq F(\mathbf{w}_1) + \nabla F(\mathbf{w}_1)^T (\mathbf{w}_2-\mathbf{w}_1) + \frac{L}{2} \|\mathbf{w}_2 - \mathbf{w}_1\|^2,\\
		&\|\nabla F(\mathbf{w}_2)-\nabla F(\mathbf{w}_1)\| \leq L \|\mathbf{w}_2 - \mathbf{w}_1\|.
	\end{align}
	
	\textbf{Assumption 3 (Gradient Variance Bound):} The local mini-batch stochastic gradient $\nabla F_n^l(\mathbf{w},\boldsymbol \xi)$ with $|\boldsymbol \xi| = B$ serves as an unbiased estimator of the actual gradient $\nabla F_n^l(\mathbf{w})$, possessing a variance that is limited as follows.
	\begin{align}
		\mathbb{E}\left\{\|\nabla F_n^l(\mathbf{w},\boldsymbol \xi) - \nabla F_n^l(\mathbf{w})\|^2\right\} \leq \frac{\sigma^2}{B}.
	\end{align}

	\textbf{Assumption 4 (Polyak-Lojasiewicz Inequality):} Let \( F^* = F(\mathbf{w}^*) \) be from problem \eqref{globalopt}. There exists a constant \( \delta \geq 0 \) for which the subsequent condition holds.
	\begin{align}
		\label{polyak}
		\|\nabla F(\mathbf{w})\|^2 \geq 2\delta \left(F(\mathbf{w}) - F^*\right).
	\end{align}
	The inequality presented in \eqref{polyak} is significantly more expansive and general than the mere assumption of convexity \cite{karimi}.

	\begin{theorem}
		Under the following conditions on the learning rate $\mu$:
		\begin{align}
			\label{conasl1}
			&1-{L^2\mu^2} \left(\tau \gamma+\frac{\tau(\tau-1)}{2}+q_1(\tau+\gamma)\max_{l}\frac{1}{N_l}\right)-  \nonumber\\&{L\mu}\left(\tau+\frac{q_1}{N}+\frac{q_2 q_1}{N}+\frac{\tau q_2 \max_l N_l}{N} \right) \geq 0,
		\end{align}
		and
		\begin{align}
			\label{conasl2}
			&1-{L^2\mu^2}\frac{\gamma(\gamma-1)}{2}-\nonumber\\&{L\mu}\gamma \left(1 +\frac{(1+q_2)q_1}{N}+\frac{q_2 \max_l N_l}{N} \right) \geq 0,
		\end{align}
		the optimality gap of {\fontfamily{lmtt}\selectfont {QHetFed}} is characterized as
		\begin{align}
			\label{op_gap}
			&\mathbb{E}\left\{F({\mathbf{w}}_{T})\right\}-F^* \leq c^T \biggl(\mathbb{E}\left\{F({\mathbf{w}}_{0})\right\}-F^*\biggr)+\frac{1-c^T}{1-c}e,
		\end{align}
		where
		\begin{align}
			c = 1-\mu (\tau+\gamma) \delta,
		\end{align}
		\begin{align}
			&e = \frac{L\mu^2}{2}\frac{\sigma^2}{B}\biggl(\frac{L \mu}{N}C(1+q_1) \tau \left[\frac{\tau-1}{2}+  \gamma\right] +{L\mu}  \frac{\gamma(\gamma-1)}{2}\nonumber\\&+\frac{1}{N} (\tau+\gamma)(1+q_2)\left(1+ {q_1}\right)\biggr) + \frac{\mu (\tau+\gamma)}{2}G^2.
		\end{align}
	\end{theorem}
	
	\begin{IEEEproof}
		See Appendix.
	\end{IEEEproof}
	
	\begin{remark}
		The term $c$ in the optimality gap indicates the speed of convergence. On the other hand, the term $e$ in the optimality gap denotes the error measure, i.e., the persistent bias post-convergence, stemming from imperfections in the learning procedure, including quantization errors, data heterogeneity, and mini-batch stochastic computations.
	\end{remark}
	
	\begin{remark}
		The maximum and minimum values of ${N_l}, \forall l$ have critical roles in determining the learning rate. Consequently, device sets of same size allow the highest learning rate.
	\end{remark}
	
	\begin{remark}
		Higher data heterogeneity $G^2$ and quantization error variances $q_i, \forall i$ lead to higher optimality gap, however, their effects are independent from each other.
	\end{remark}
	
	In the subsequent corollary, we evaluate how \(q_1\) influences the impact of \(\tau\) and \(\gamma\) on performance.
	\begin{corollary}
		Given a constant sum for \(\tau+\gamma\), if \(q_1 < \frac{N}{C} - 1\), then a higher \(\tau\) leads to a reduced optimality gap. On the other hand, if \(q_1 > \frac{N}{C} - 1\), decreasing \(\tau\) results in a smaller optimality gap.
	\end{corollary}
	\begin{IEEEproof}
		Only the following term of $e$ in the optimality gap \eqref{op_gap} is not a function of $\tau+\gamma$, which we denote by $\beta$.
		\begin{align}
			&\frac{C}{N}(1+q_1) \tau \left[\frac{\tau-1}{2}+  \gamma\right] +\frac{\gamma(\gamma-1)}{2} = \frac{C}{N}(1+q_1) \tau \times \nonumber\\& \left[\frac{\tau-1}{2}+  \beta - \tau \right] +\frac{(\beta- \tau)(\beta-\tau-1)}{2}=\nonumber\\& \frac{1}{2} \left[1-\frac{C}{N}(1+q_1)\right] \tau^2 - \left(\beta - \frac{1}{2}\right)\left[1-\frac{C}{N}(1+q_1)\right] \tau  \nonumber\\&=\left[1-\frac{C}{N}(1+q_1)\right]\left[\frac{1}{2}\tau^2 - \left(\beta - \frac{1}{2}\right)\tau\right].
		\end{align} 
		Given that \(\beta - \frac{1}{2} = \tau+\gamma - \frac{1}{2} > \tau\), it follows that when the scaling factor \(1-\frac{C}{N}(1+q_1) > 0\), an increase in \(\tau\) results in a reduction of \(\frac{1}{2}\tau^2 - \left(\beta - \frac{1}{2}\right)\tau\), thereby a reduction of the optimality gap.
	\end{IEEEproof}
	\begin{remark}
		The variance of the quantization error $q$, decreases with the number of quantization levels. Therefore, the results in Corollary 1 mean that under a high number of quantization levels, it is better to have more intra-set iterations, while under lower number of quantization levels it is better to transmit less and increase the number of gradient descent steps within each global iteration. This highlights the importance of co-designing the learning algorithm and the transmission scheme to achieve optimal performance.
	\end{remark}

	In the special case of devices with low computational capabilities, it is necessary to minimize local computations by limiting it to just a single step of local training. That is, $\gamma = 1$ and $\tau$ is arbitrary. In this case, the edge servers are aware of the model parameters within their sets at the end of the intra-set iterations, which allows the following simplified hierarchical algorithm: 
	The edge servers and the devices perform one-step gradient descent as (6) for $\tau+1$ intra-set iterations. Then, the edge servers forward the model parameters to the cloud server, and cloud aggregation is performed according to (10). The specialized optimality gap is given in the next corollary.
	\begin{corollary}
		Under $\gamma = 1$ and the following condition on the learning rate $\mu$:
		\begin{align}
			\label{conalt}
			&1-{L^2\mu^2} \left(\tau +\frac{\tau(\tau-1)}{2}+q_1(\tau+1)\max_{l}\frac{1}{N_l}\right)  -\nonumber\\&{L\mu}(1+q_2)\Bigl(\frac{\tau \max_l N_l}{N}+\frac{q_1}{N}\Bigr)\geq 0,
		\end{align}
		the error term in the optimality gap is as
		\begin{align}
			&e = \frac{L\mu^2}{2}\frac{\sigma^2}{B}\biggl(\frac{L \mu}{N}C(1+q_1)  \frac{(\tau+1)\tau}{2} +\nonumber\\&\frac{1}{N} (\tau+1)(1+q_2)\left(1+ {q_1}\right)\biggr) + \frac{\mu (\tau+1)}{2}G^2.
		\end{align}
		
	\end{corollary}
	\begin{IEEEproof}
		This is achieved by setting $\gamma = 1$ and the fact that the condition from \eqref{conasl2}, specifically $1-{L\mu} \left(1 +\frac{(1+q_2)q_1}{N}+\frac{q_2 \max_l N_l}{N} \right) \geq 0$, holds true when the condition \eqref{conalt} is met. 
	\end{IEEEproof}
	
	{\fontfamily{lmtt}\selectfont {QHetFed}}, with its periodic aggregation of gradients and model parameters, introduces a novel concept even for standard FL systems that lack a hierarchical structure, i.e., single-cell FL. It is expected to provide resilience against data heterogeneity, as the hierarchical scheme. In that case, the edge server and the cloud server are the same physical units, and the simplified optimality gap is as follows.
	\begin{corollary}
		When $C = 1$ and $q_2 = 0$, under the following conditions on the learning rate $\mu$:
		\begin{align}
			&1-{L^2\mu^2} \left(\tau \gamma+\frac{\tau(\tau-1)}{2}+\frac{q_1(\tau+\gamma)}{N}\right)  -\nonumber\\&{L\mu}\left(\tau+\frac{q_1}{N} \right) \geq 0,
		\end{align}
		and
		\begin{align}
			1-{L^2\mu^2}\frac{\gamma(\gamma-1)}{2}-{L\mu}\gamma \left(1 +\frac{q_1}{N}\right) \geq 0,
		\end{align}
		the error term in the optimality gap is as
		\begin{align}
			&e = \frac{L\mu^2}{2}\frac{\sigma^2}{B}\biggl(\frac{L \mu}{N}(1+q_1) \tau \left[\frac{\tau-1}{2}+  \gamma\right] +{L\mu}  \frac{\gamma(\gamma-1)}{2}\nonumber\\&+\frac{1}{N} (\tau+\gamma)\left(1+ {q_1}\right)\biggr)+ \frac{\mu (\tau+\gamma)}{2}G^2.
		\end{align}
	\end{corollary}
	
	\section{Comparison with the Conventional Hierarchical Scheme}
	The primary hierarchical FL algorithm integrating quantization, named {\fontfamily{lmtt}\selectfont {{Hier-Local-QSGD}}}, is introduced in \cite{letaief} and detailed in Algorithm 2. This algorithm, {\fontfamily{lmtt}\selectfont {{Hier-Local-QSGD}}}, conducts model parameter aggregation at both hierarchical levels. Its key distinction from {\fontfamily{lmtt}\selectfont {QHetFed}} lies in the intra-set update phase, denoted by *, which now includes successive local steps. We present the analytical comparison of the two approaches, the proposed {\fontfamily{lmtt}\selectfont {{QHetFed}}} and {\fontfamily{lmtt}\selectfont {{Hier-Local-QSGD}}}. First, the optimality gap of {\fontfamily{lmtt}\selectfont {{Hier-Local-QSGD}}} is derived, under the conditions described in Subsection II. B. Based on this, subsequent remarks compare the learning performance of the schemes.
	
	\begin{lemma}
		Under the following single condition on $\mu$:
		\begin{align}
			&1- L^2 \mu^2 \left(\frac{\gamma(\gamma-1)}{2}+\gamma \tau \left(\frac{\tau(\tau-1)}{2}+q_1\tau\right)\right) -\nonumber\\& L\mu (1+q_2)\left(\gamma\tau+ \frac{q_1\gamma}{N}\right) \geq 0,
		\end{align}
		the optimality gap of {\fontfamily{lmtt}\selectfont {{Hier-Local-QSGD}}} is characterized as
		\begin{align}
			&\mathbb{E}\left\{F({\mathbf{w}}_{T})\right\}-F^* \leq {\bar c}^{T} \biggl(\mathbb{E}\left\{F({\mathbf{w}}_{0})\right\}-F^*\biggr)+\frac{1-{\bar c}^T}{1-{\bar c}}{\bar e},
		\end{align}
		where
		\begin{align}
			{\bar c} = 1-\mu \tau\gamma \delta,
		\end{align}
		\begin{align}
			&{\bar e} = \frac{L\mu^2}{2}\frac{\sigma^2}{B}\biggl( \frac{{L\mu}}{N}C(1+q_1) \frac{\gamma^2 \tau (\tau-1)}{2} +{L\mu}\frac{\tau \gamma(\gamma-1)}{2}\nonumber\\&+\frac{1}{N} \tau \gamma (1+q_2)\left(1+ {q_1}\right)\biggr)+\frac{\mu \tau \gamma}{2}G^2.
		\end{align}	
	\end{lemma}
	\begin{IEEEproof}
		Theorem 1 in \cite{letaief} presents a convergence rate analysis that is limited to i.i.d. data and excludes a term for data heterogeneity. By adopting the methodology outlined in \cite{letaief} and making necessary adjustments to incorporate data heterogeneity according to our approach in Appendix, we can derive the convergence rate of {\fontfamily{lmtt}\selectfont {{Hier-Local-QSGD}}}. Subsequently, by implementing the final step described in \eqref{proof_final_step} in Appendix, the corresponding optimality gap can be determined. While new, the detailed proof is omitted here.
	\end{IEEEproof}
	\begin{remark}
		For {\fontfamily{lmtt}\selectfont {{Hier-Local-QSGD}}}, the convergence speed is scaled by $\gamma \tau$, whereas in {\fontfamily{lmtt}\selectfont {QHetFed}}, it is boosted by $\tau+\gamma$. This distinction arises because {\fontfamily{lmtt}\selectfont {{Hier-Local-QSGD}}} incorporates 
		$\gamma$ local steps in each intra-set iteration. In contrast, {\fontfamily{lmtt}\selectfont {QHetFed}} has a single step for local updates in the intra-set iterations. Thus, it is evident that {\fontfamily{lmtt}\selectfont {{Hier-Local-QSGD}}} is faster than {\fontfamily{lmtt}\selectfont {QHetFed}} in achieving convergence.
	\end{remark}

	Although the convergence speed is crucial for ensuring low latency learning, in many learning systems, the error measure is prioritized over convergence speed. This is because the primary goal of any learning task is accuracy. 
	
	The difference in the error terms for the two methods is as 
	\begin{align}
		\label{diff}
		&\Delta = \bar{e} - e = \frac{L\mu^2}{2}\frac{\sigma^2}{B}\biggl( \frac{{L\mu}}{N}C(1+q_1) \frac{\gamma^2 \tau (\tau-1)}{2} +\nonumber\\&{L\mu}\frac{\tau \gamma(\gamma-1)}{2}+\frac{1}{N} \tau \gamma (1+q_2)\left(1+ {q_1}\right)\biggr)+\frac{\mu \tau \gamma}{2}G^2 -\nonumber\\& \frac{L\mu^2}{2}\frac{\sigma^2}{B}\biggl(\frac{L \mu}{N}C(1+q_1) \tau \left[\frac{\tau-1}{2}+  \gamma\right] +{L\mu}  \frac{\gamma(\gamma-1)}{2}+\nonumber\\&\frac{1}{N} (\tau+\gamma)(1+q_2)\left(1+ {q_1}\right)\biggr) - \frac{\mu (\tau+\gamma)}{2}G^2 = \frac{L\mu^2}{2}\frac{\sigma^2}{B}\nonumber\\&\biggl( \frac{{L\mu}}{N}C(1+q_1) \left(\frac{\gamma^2 \tau (\tau-1)}{2} - \frac{ \tau (\tau-1)}{2} - \tau \gamma\right) +\nonumber\\
		& \frac{1}{N}  (1+q_2)\left(1+ {q_1}\right)(\tau \gamma - \tau-\gamma)+{L\mu}\frac{(\tau-1) \gamma(\gamma-1)}{2}\biggr) \nonumber\\
		& +\frac{\mu}{2}G^2 (\tau\gamma  - \tau -\gamma),
	\end{align}
	which is positive, denoting a consistently higher error for {\fontfamily{lmtt}\selectfont {Hier-Local-QSGD}} in comparison to {\fontfamily{lmtt}\selectfont {QHetFed}}. This increase comprises four different parts:
	
	\textit{i) Increase because of quantization layer 1 as} 
	\begin{align}
		&\Delta^{Q_1} = \frac{L^2\mu^3}{2N}\frac{\sigma^2}{B} C(1+q_1) \left(\frac{(\gamma^2 - 1) \tau (\tau-1)}{2} - \tau \gamma\right).
	\end{align}
	
	\textit{ii) Increase because of quantization layer 2 as}
	\begin{align}
		&\Delta^{Q_2} = \frac{L\mu^2}{2N}\frac{\sigma^2}{B}  (1+q_2)\left(1+ {q_1}\right)(\tau \gamma - \tau-\gamma).
	\end{align}
	
	\textit{iii) Increase because of stochastic local computations as}
	\begin{align}
		&\Delta^{\text{local-comp}} = \frac{L^2\mu^3}{2}\frac{\sigma^2}{B} \frac{(\tau-1) \gamma(\gamma-1)}{2}.
	\end{align}
	
	\textit{iv) Increase because of data heterogeneity as}
	\begin{align}
		&\Delta^{\text{het}} = \frac{\mu}{2}G^2 (\tau\gamma  - \tau -\gamma).
	\end{align}
	
	\begin{remark}
		The increase term $\Delta$ intensifies when there is an increase in any of the parameters such as $\gamma$, $\tau$, $q_1$, $q_2$, $C$, $\frac{\sigma^2}{B}$, or $G^2$. This underlines the effectiveness of {\fontfamily{lmtt}\selectfont {QHetFed}} particularly in scenarios where these parameters have sufficiently high values.
	\end{remark}
	\begin{remark}
		The {\fontfamily{lmtt}\selectfont {{Hier-Local-QSGD}}} outperforms {\fontfamily{lmtt}\selectfont {QHetFed}} under i.i.d. data and low quantization errors.	
	\end{remark}

	\begin{algorithm}
		\small
		\caption{{\fontfamily{lmtt}\selectfont {{Hier-Local-QSGD}}} algorithm}
		\begin{algorithmic}
			\vspace{0pt}
			\State Initialize the global model $\mathbf{w}_0$\
			\vspace{0pt}
			\State \textbf{for} inter-set iteration $t=1,...,T$ \textbf{do}
			\vspace{0pt}
			\State \hspace{10pt}Each device updates its model by $\mathbf{w}_{n,1,0,t}^l=\mathbf{w}_{t}$ 
			\vspace{0pt}
			\State \hspace{10pt}\textbf{for} intra-set iteration $i=1,...,\tau$ \textbf{do}
			\vspace{0pt}
			\State \hspace{15.7pt}*Each device updates its local model as $\small{
				\mathbf{w}_{n,i,j,t}^{l} = \mathbf{w}_{n,i,j-1,t}^{l} - \mu}\small{\nabla F_{n}^{l}(\mathbf{w}_{n,i,j-1,t}^{l},\boldsymbol\xi_{n,i,j-1,t}^l), \ j \leq \gamma}$
			\vspace{0pt}
			\State \hspace{20pt}Each edge server obtains its intra-set model vector from $\mathbf{w}_{i+1,t}^{l} = \mathbf{w}_{i,t}^{l} +\frac{1}{N_l}\sum_{n}^{} Q_1(\mathbf{w}_{n,i,\gamma, t}^{l}-\mathbf{w}_{n,i,0, t}^{l})$
			\vspace{0pt}
			\State \hspace{20pt}Each device updates its model by $\mathbf{w}_{n,i+1,0,t}^l=\mathbf{w}_{i+1,t}^l$
			\vspace{10pt}
			\State \hspace{10pt}Each edge server obtains its intra-set model from $\mathbf{w}_{t+1}^{l} = \mathbf{w}_{t}^{l}+ \frac{1}{N_l}\sum_{n}^{} Q_1(\mathbf{w}_{n,\tau,\gamma,t}^{l}-\mathbf{w}_{n,\tau,0,t}^{l})$
			\vspace{0pt}
			\State \hspace{10pt}Cloud server obtains global model from $\mathbf{w}_{t+1} = \mathbf{w}_t +\frac{1}{N_l} \sum_{l}^{} N_l Q_2(\mathbf{w}_{t+1}^{l}-\mathbf{w}_t)$
		\end{algorithmic}
	\end{algorithm}
	
	\section{System Optimization}
	The values of the number of intra-set iterations and gradient descent steps, $\tau$ and $\gamma$, can be chosen to minimize the optimality gap, as the most beneficial metric for achieving the highest learning accuracy. However, due to its complex form, we choose to consider a more tractable alternative metric which is based on the convergence rate of {\fontfamily{lmtt}\selectfont {QHetFed}}, given in the next lemma. The convergence rate has been extensively utilized for optimizations in other FL research works, e.g., \cite{conv1, conv2, conv3, conv4}.
	\begin{lemma}
		Under the conditions \eqref{conasl1} and \eqref{conasl2} on $\mu$, the convergence rate of {\fontfamily{lmtt}\selectfont {QHetFed}} is characterized as
		\begin{align}
			\label{con_rate}
			&\frac{1}{T} \sum_{t=0}^{T-1}\mathbb{E}\left\{\| \nabla F(\mathbf{w}_t)\|^2\right\} \leq \frac{2(F(\mathbf{w}_0)-F^*)}{\mu (\tau+\gamma)T}+\nonumber\\& 
			\frac{L^2\mu^2}{2}\frac{\sigma^2}{B}\left( \frac{C}{N}(1+q_1) \tau \left(1+\frac{\gamma-1}{\tau+\gamma}\right)+\frac{{\gamma(\gamma-1)}}{\tau+\gamma}\right)+\nonumber\\&{L\mu}\frac{\sigma^2}{B}\frac{1}{N} (1+q_2)\left(1+ {q_1}\right)+G^2.
		\end{align}
	\end{lemma} 
	\begin{IEEEproof}
		After performing a telescoping sum over \eqref{lipexp2} in Appendix for the global iterations $t \in \left\{0,\cdots,T-1\right\}$ and using the fact $\mathbb{E}\left\{F(\mathbf{w}_{T})\right\}\geq F^{*}$, we reach the conclusion of the proof.	
	\end{IEEEproof}
	
	The goal of the parameter optimization then would be to minimize \eqref{con_rate}, considering that the learning process can run until a deadline $T_\text{d}$ for delay-constrained applications. For {\fontfamily{lmtt}\selectfont {QHetFed}}, $T_\text{d}$ needs to cover the computation and communication times as  
	\begin{align}
		\label{runtime1}
		T_\text{d} = T \times T_\text{di}(\tau, \gamma), 
	\end{align}
	where
	\begin{align}
		\label{delay_it}
		T_\text{di}(\tau,\gamma) = (\tau+\gamma) t_\text{CP}+\tau t_\text{DE} + t_\text{EC}, 
	\end{align}
	is the delay per global iteration. In \eqref{delay_it}, $t_\text{CP}$ represents the computation time at each device, while $t_\text{DE}$ and $t_\text{EC}$ denote the communication times between each device and its respective edge server and between each edge server and the cloud server, respectively, with $t_\text{EC} \gg t_\text{DE}$. From \cite{letaief}, these parameters can be obtained as
	$
	t_\text{CP} = \frac{cD}{f},
	$
	$
	t_\text{DE} = \frac{d_\text{b}}{B\log_2\left(1+\frac{hp}{N_0}\right)},
	$
	where $c$ is the number of CPU cycles to
	execute one sample of data, $f$ is the CPU cycle frequency, $D$ is the number of data bits involved in one local iteration, $d_\text{b}$ is the model size in bits, $B$ is the channel bandwidth, $h$ is the channel gain, $p$ is the transmission power, and $N_0$ is the noise power. 
	
	The first term of the right-hand side (RHS) of \eqref{con_rate}, i.e., $\frac{2(F(\mathbf{w}_0)-F^*)}{\mu (\tau+\gamma)T}$, complicates the accurate assessment of the RHS. This complexity arises because determining the values of $F^{*}$, $L$, and $\sigma^2$ in \eqref{con_rate} requires prior statistical knowledge of the local learning models and data statistics, which is unavailable in many applications. Therefore, we suggest to select the second term as our optimization objective. This term represents the error in the $l_2$ norm of the global gradient, which is a key factor in progressing towards convergence. Moreover, $T_\text{d}$ in \eqref{runtime1} incorporates $T(\tau+\gamma)$, and thus the convergence rate according to the disregarded first term of \eqref{con_rate}.
	
	Based on this, we suggest to select the value of $\tau$ and $\gamma$ by solving the following optimization problem.
	\begin{align}
		\label{constraint}
		\min_{\tau, \gamma} \frac{C}{N}(1+q_1) \tau \left(1+\frac{\gamma-1}{\tau+\gamma}\right)+\frac{{\gamma(\gamma-1)}}{\tau+\gamma},
	\end{align}
	subject to \eqref{runtime1}. From \eqref{runtime1} and \eqref{delay_it}, we have
	\begin{align}
		\label{forgamma}
		&\tau + \gamma = \frac{T_\text{d}}{Tt_\text{CP}} - \frac{t_\text{DE}}{t_\text{CP}} \tau - \frac{t_\text{EC}}{t_\text{CP}},\nonumber\\&
		\gamma = \frac{T_\text{d}}{Tt_\text{CP}} - \left(1+\frac{t_\text{DE}}{t_\text{CP}}\right) \tau - \frac{t_\text{EC}}{t_\text{CP}},
	\end{align}
	whereby the optimization problem becomes
	\begin{align}
		&\min_{\tau} \Biggl\{\frac{C}{N}(1+q_1) \tau \left(1+\frac{\gamma-1}{\tau+\gamma}\right)+\frac{{\gamma(\gamma-1)}}{\tau+\gamma} = \nonumber\\&\frac{C}{N}(1+q_1) \tau \left(2-\frac{1+\tau}{\frac{T_\text{d}}{Tt_\text{CP}} - \frac{t_\text{DE}}{t_\text{CP}} \tau - \frac{t_\text{EC}}{t_\text{CP}}}\right)+\nonumber\\&\left(\frac{T_\text{d}}{Tt_\text{CP}} - \left(1+\frac{t_\text{DE}}{t_\text{CP}}\right) \tau - \frac{t_\text{EC}}{t_\text{CP}}\right)
		\left(1-\frac{1+\tau}{\frac{T_\text{d}}{Tt_\text{CP}} - \frac{t_\text{DE}}{t_\text{CP}} \tau - \frac{t_\text{EC}}{t_\text{CP}}}\right)\nonumber\\
		&= \left(1-\frac{1+\tau}{\frac{T_\text{d}}{Tt_\text{CP}} - \frac{t_\text{DE}}{t_\text{CP}} \tau - \frac{t_\text{EC}}{t_\text{CP}}}\right) \biggl(\left(\frac{C}{N}(1+q_1)   - 1-\frac{t_\text{DE}}{t_\text{CP}} \right)\tau \nonumber\\&+ \frac{T_\text{d}}{Tt_\text{CP}} - \frac{t_\text{EC}}{t_\text{CP}}\biggr)+\frac{C}{N}(1+q_1) \tau \triangleq J(\tau)\Biggr\}.
	\end{align}
	This problem can be solved by taking derivative from its objective with respect to $\tau$ as
	\begin{align}
		&\left(\frac{C}{N}(1+q_1)   - 1-\frac{t_\text{DE}}{t_\text{CP}} \right)\left(1-\frac{1+\tau}{\frac{T_\text{d}}{Tt_\text{CP}} - \frac{t_\text{DE}}{t_\text{CP}} \tau - \frac{t_\text{EC}}{t_\text{CP}}}\right)+\nonumber\\&\frac{-\frac{T_\text{d}}{Tt_\text{CP}}  + \frac{t_\text{EC}}{t_\text{CP}}-\frac{t_\text{DE}}{t_\text{CP}}}{\left(\frac{T_\text{d}}{Tt_\text{CP}} - \frac{t_\text{DE}}{t_\text{CP}} \tau - \frac{t_\text{EC}}{t_\text{CP}}\right)^2} \frac{}{}\biggl(\left(\frac{C}{N}(1+q_1)   - 1-\frac{t_\text{DE}}{t_\text{CP}} \right)\tau+\nonumber\\
		& \frac{T_\text{d}}{Tt_\text{CP}} - \frac{t_\text{EC}}{t_\text{CP}}\biggr)+\frac{C}{N}(1+q_1) = 0,
	\end{align}
	which is equal to $a_0\tau^2+b_0\tau+c_0 = 0$, where
	\begin{align}
		&a_0 = \left(\frac{C}{N}(1+q_1)   - 1-\frac{t_\text{DE}}{t_\text{CP}} \right)\left(\frac{t_\text{DE}^2}{t_\text{CP}^2}+\frac{t_\text{DE}}{t_\text{CP}}\right)\nonumber\\&+\frac{C}{N}(1+q_1)\frac{t_\text{DE}^2}{t_\text{CP}^2},
	\end{align}
	\begin{align}
		&b_0 = \left(\frac{C}{N}(1+q_1)   - 1-\frac{t_\text{DE}}{t_\text{CP}} \right)\biggl(\frac{t_\text{DE}}{t_\text{CP}}+\frac{t_\text{EC}}{t_\text{CP}}-\frac{T_\text{d}}{Tt_\text{CP}}-2\times\nonumber\\&\left(\frac{T_\text{d}}{Tt_\text{CP}} - \frac{t_\text{EC}}{t_\text{CP}}\right)\frac{t_\text{DE}}{t_\text{CP}} \biggr)+\left({-\frac{T_\text{d}}{Tt_\text{CP}}  + \frac{t_\text{EC}}{t_\text{CP}}-\frac{t_\text{DE}}{t_\text{CP}}} \right)\biggl(\frac{C}{N}\times\nonumber\\&(1+q_1)   - 1-\frac{t_\text{DE}}{t_\text{CP}} \biggr)-2\frac{C}{N}(1+q_1)\left(\frac{T_\text{d}}{Tt_\text{CP}} - \frac{t_\text{EC}}{t_\text{CP}}\right)\frac{t_\text{DE}}{t_\text{CP}},
	\end{align}
	\begin{align}
		&c_0 = \left(\frac{C}{N}(1+q_1)   - 1-\frac{t_\text{DE}}{t_\text{CP}} \right)\biggl(\left(\frac{T_\text{d}}{Tt_\text{CP}} - \frac{t_\text{EC}}{t_\text{CP}}\right)^2- \frac{T_\text{d}}{Tt_\text{CP}} \nonumber\\&+ \frac{t_\text{EC}}{t_\text{CP}}\biggr)+\left({-\frac{T_\text{d}}{Tt_\text{CP}}  + \frac{t_\text{EC}}{t_\text{CP}}-\frac{t_\text{DE}}{t_\text{CP}}} \right)\left( \frac{T_\text{d}}{Tt_\text{CP}} - \frac{t_\text{EC}}{t_\text{CP}}\right)+\nonumber\\&\frac{C}{N}(1+q_1)\left(\frac{T_\text{d}}{Tt_\text{CP}} - \frac{t_\text{EC}}{t_\text{CP}}\right)^2.
	\end{align}
	Thus, the optimum value of $\tau$ is $\tau_\text{opt} = \arg \min_{\left\{1, \frac{-b_0\pm\sqrt{b_0^2-4a_0c_0}}{2a_0}\right\}} J(\tau)$. Then, the optimum value of $\gamma$ is obtained from \eqref{forgamma} as $\gamma_\text{opt} = \frac{T_\text{d}}{Tt_\text{CP}} - \left(1+\frac{t_\text{DE}}{t_\text{CP}}\right) \tau_\text{opt} - \frac{t_\text{EC}}{t_\text{CP}}$.
	\section{Experimental Results}
	We consider a hierarchical network with three sets, performing image classification task. The network and learning parameters are given in Table I. We take into account that communication between edge servers and the cloud server typically utilizes high bandwidth backhaul links, and thus $s_2 \gg s_1$. CIFAR-10 \footnote{The CIFAR-10 is a widely-used standard dataset in the field of machine learning and computer vision. It comprises 60000 colored images, divided into ten classes with 6000 images each. These images are relatively complex, featuring varied subjects such as animals and vehicles, making CIFAR-10 a challenging dataset for evaluating learning algorithms \cite{cifar100}.} is utilized for the image classification task. We have constructed the classifier using a Convolutional Neural Network (CNN). This CNN consists of four $3 \times 3$ convolution layers with ReLU activation (the first two with 32 channels,
	the second two with 64), each two followed by a $2 \times 2$
	max pooling; a fully connected layer with $128$ units and ReLU
	activation; and a final softmax output layer. Both i.i.d. and non-i.i.d. distributions of dataset samples among devices are considered. For the non-i.i.d. setting, each device contains samples exclusively from two randomly selected classes out of the ten available classes in CIFAR-10. The sample count differs from one device to another, following a uniform distribution within the range $[500, 1500]$. Performance is measured by evaluating the learning accuracy on the test dataset over the global iteration count, denoted by $t$. The final performance results are obtained by averaging the outcomes from 20 different runs.

	\begin{table}
		\caption {Algorithm Parameters} 
		\vspace{-8pt}
		\begin{center}
			\resizebox{6.4cm}{!} {
				\begin{tabular}{| l | l | l | l | l | l | l | l | l | l | l | l |}
					
					\hline
					\hline
					{$C$}&{$N_l, \forall l$}&{$\tau$}& $\gamma$&$\mu$ &$B$& $s_1$ &$s_2$ \\ \hline
					$3$& $20$ &$12$&$3$&$0.01$& $100$& $4$&$10$  \\ \hline	
					\hline
			\end{tabular}}
		\end{center}
		\vspace{-8pt}
	\end{table}
	
	\begin{figure}[tb!]
		\vspace{0pt}
		\centering
		\includegraphics[width =2.5in]{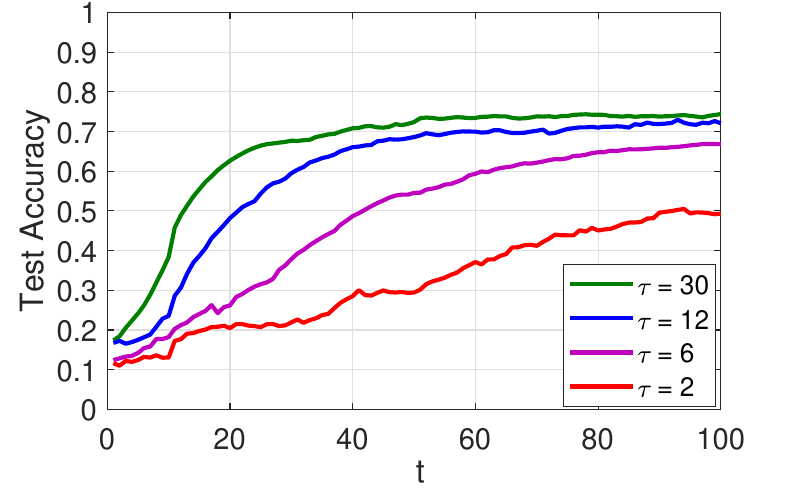} 
		\vspace{-3pt}
		\caption{Test accuracy as a function of global iterations (i.i.d.)}
		\vspace{-5pt}
	\end{figure}
	
	\begin{figure}[tb!]
		\vspace{0pt}
		\centering
		\includegraphics[width =2.5in]{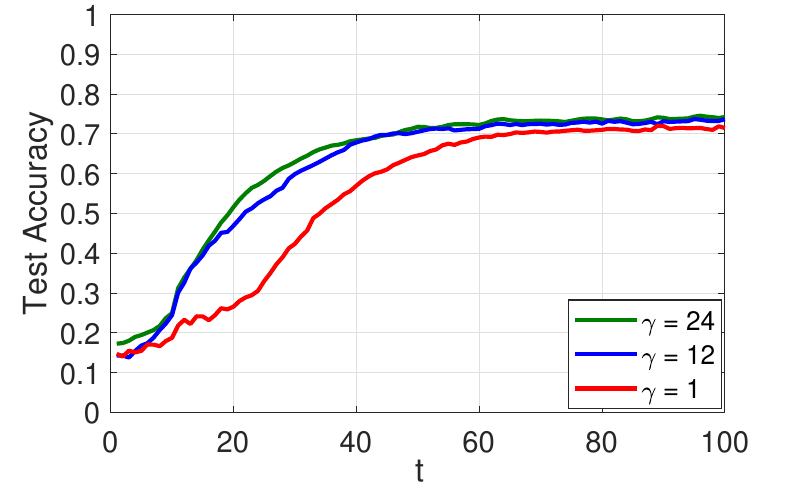} 
		\vspace{-3pt}
		\caption{Test accuracy as a function of global iterations (i.i.d.)}
		\vspace{-5pt}
	\end{figure}
	
	\begin{figure}[tb!]
		\vspace{0pt}
		\centering
		\includegraphics[width =2.5in]{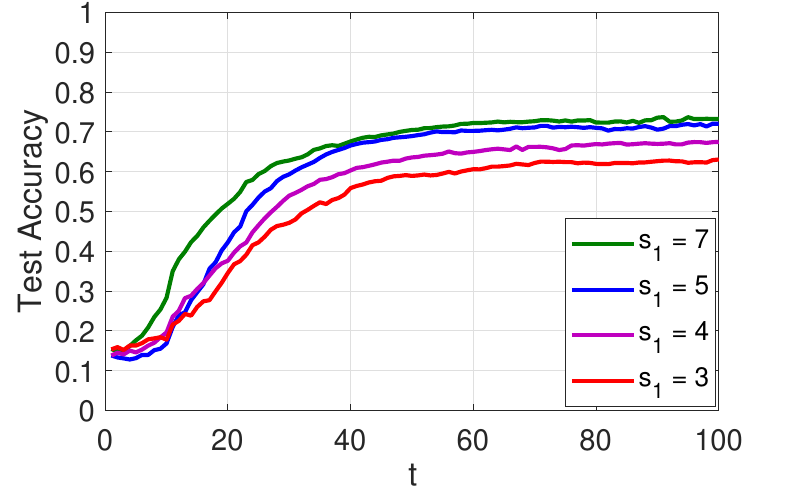} 
		\vspace{-3pt}
		\caption{Test accuracy as a function of global iterations (non-i.i.d.)}
		\vspace{-5pt}
	\end{figure}
	
	\begin{figure}[tb!]
		\vspace{0pt}
		\centering
		\includegraphics[width =2.5in]{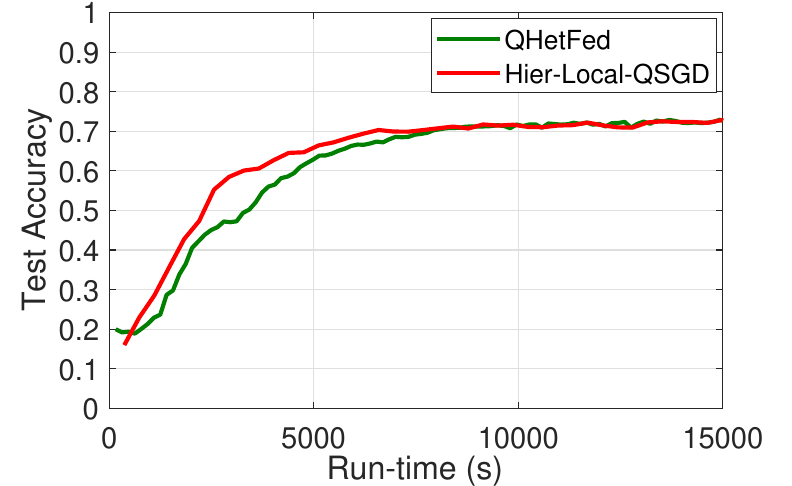} 
		\vspace{-3pt}
		\caption{Test accuracy as a function of run-time (i.i.d.)}
		\vspace{-5pt}
	\end{figure}
	
	\begin{figure}[tb!]
		\vspace{0pt}
		\centering
		\includegraphics[width =2.5in]{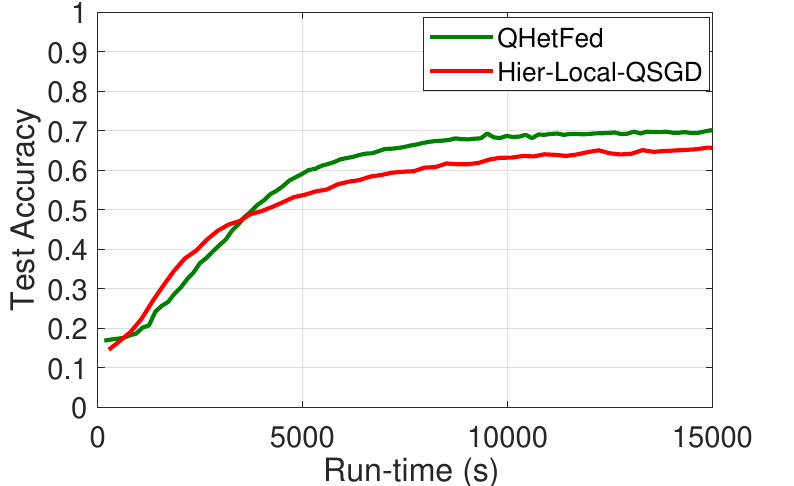} 
		\vspace{-3pt}
		\caption{Test accuracy as a function of run-time (mixed)}
		\vspace{-5pt}
	\end{figure}

	\begin{figure}[tb!]
		\vspace{0pt}
		\centering
		\includegraphics[width =2.5in]{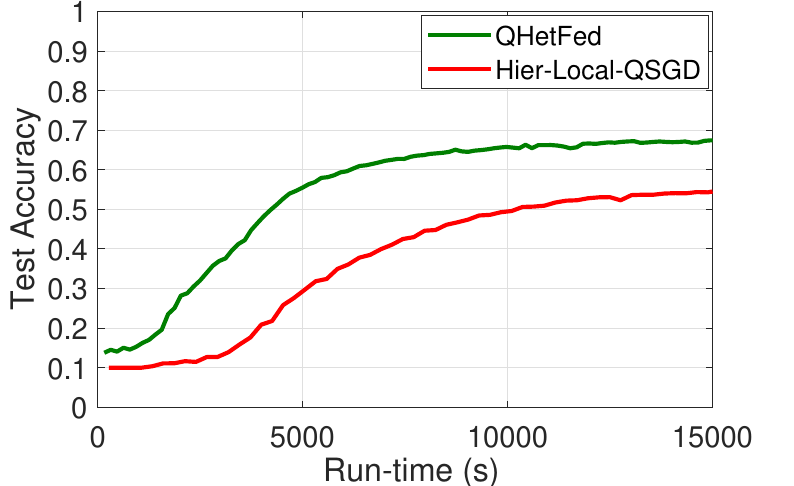} 
		\vspace{-3pt}
		\caption{Test accuracy as a function of run-time (non-i.i.d.1)}
		\vspace{-5pt}
	\end{figure}
	
	\begin{figure}[tb!]
		\vspace{0pt}
		\centering
		\includegraphics[width =2.5in]{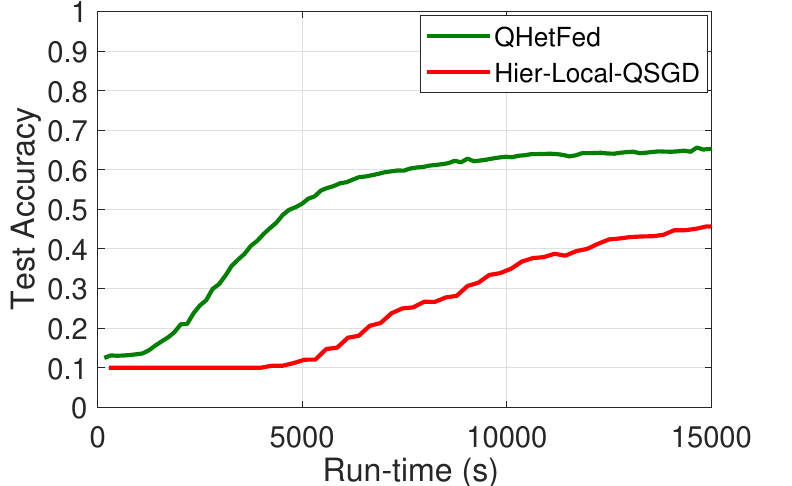} 
		\vspace{-3pt}
		\caption{Test accuracy as a function of run-time (non-i.i.d.2)}
		\vspace{-5pt}
	\end{figure}
	
	\begin{figure}[tb!]
		\vspace{0pt}
		\centering
		\includegraphics[width =2.5in]{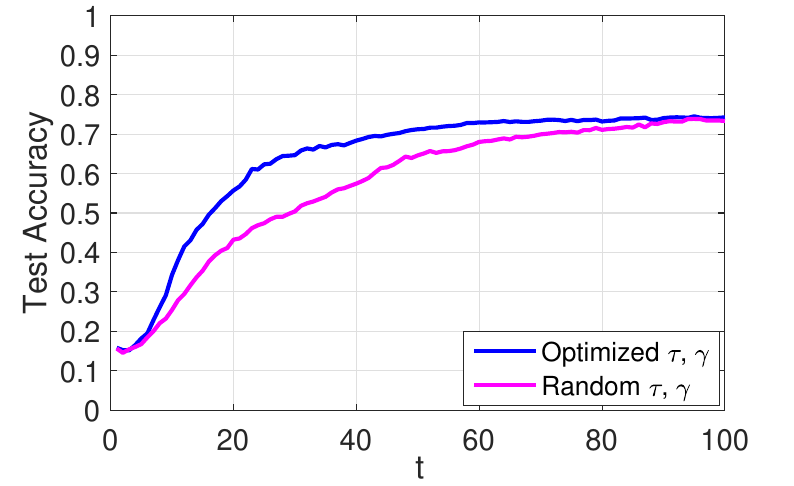} 
		\vspace{-3pt}
		\caption{Test accuracy as a function of global iterations (non-i.i.d.1)}
		\vspace{-5pt}
	\end{figure}
	
	Fig. 2 displays the accuracy for varying numbers of intra-set iterations $\tau$ in the i.i.d. setting. It is observed that increasing $\tau$ or $t$ boosts the learning performance. However, the margin of improvement narrows at higher values of $\tau$ or $t$. Furthermore, elevating $\tau$ leads to faster convergence in terms of $t$.
	
	Fig. 3 presents the accuracy across various number of local iterations $\gamma$ within the i.i.d. setting. There is an enhancement in performance as $\gamma$ increases, though the improvement gap diminishes at higher levels of $\gamma$. This highlights the vital importance of conducting multiple-step local learning at the conclusion of each inter-set iteration in our approach.
	
	In Fig. 4, the impact of varying the number of quantization levels $s_1$ in the quantization function $Q_1$ is explored in the non-i.i.d. setting. It is observed that an increase in $s_1$ results in enhanced performance, attributable to more accurate transmission. Nonetheless, the improvement gap narrows with higher values of $s_1$. Additionally, using a very low number of quantization levels, such as $s_1 = 3$, yields stable and acceptable performance. These suggest that a minimum level of $s_1$ is adequate for achieving satisfactory performance, highlighting the robustness of our scheme against the adverse effects of quantization. 
	
	\begin{table}[t]
		\centering
		\caption{Test accuracy at $t=100$.}
		\label{your_label_here}
		\vspace{-2pt}
		\begin{tabular}{|c|c|c|c|c|}
			\hline
			& \multicolumn{2}{c|}{$s_1 = 3, q_1 = 149.3$} & \multicolumn{2}{c|}{$s_1 = 7, q_1 = 11.9$} \\ \hline
			$(\tau,\gamma)$ & (15,5) & (10,10) & (15,5) & (10,10)\\ \hline
			i.i.d. case & 0.7234 & 0.7579 & 0.8213 & 0.8108 \\ \hline
			non-i.i.d. case & 0.6742 & 0.6835 & 0.7477 & 0.7290 \\ \hline
		\end{tabular}
		\vspace{-2pt}
		
	\end{table}
	
	Table II presents the accuracy at $t = 100$ for various $\tau$, $\gamma$, adhering to the constraint $\tau + \gamma = 20$, and $q_1$ \footnote{In our work, the quantization error variance $q$ is measured numerically.} corresponding to $s_1 = \left\{3, 7\right\}$, in both i.i.d. and non-i.i.d. settings. It is noted that a higher $\gamma$ enhances performance when $q_1$ is high. Conversely, an increased $\tau$ leads to improved performance when $q_1$ is low. It justifies Corollary 1 and \textit{Remark 4}.

	\begin{table}
		\caption {Run-time parameters} 
		\vspace{-8pt}
		\begin{center}
			\resizebox{8.5cm}{!} {
				\begin{tabular}{| l | l | l | l | l | l | l | l | l | l | l | l}
					
					\hline
					\hline
					{$B$}&{$p$}&{$N_0$}& $c$&$h$ &$f$&$t_\text{EC}$ \\ \hline
					$1\ \text{MHz}$& $0.5\ \text{W}$ &$10^{-10}\ \text{W}$&$20\ \text{cycles/bit}$&$10^{-8}$& $1\ \text{GHz}$&$10 t_\text{DE}$  \\ \hline	
					\hline
			\end{tabular}}
		\end{center}
		\vspace{-17pt}
	\end{table}
	
	In Figs 5-8, we compare the learning performance of {\fontfamily{lmtt}\selectfont QHetFed} with the conventional hierarchical FL ({\fontfamily{lmtt}\selectfont Hier-Local-QSGD}) from \cite{letaief} across four different data distribution scenarios: i.i.d., mixed, and two non-i.i.d. settings. To ensure a fair comparison, the accuracy is plotted against the algorithm runtime, given by $tT_\text{di}(\tau, \gamma), \forall t$ for {\fontfamily{lmtt}\selectfont
		QHetFed}. For {\fontfamily{lmtt}\selectfont Hier-Local-QSGD}, the runtime is specified based on the reasoning in \eqref{delay_it} and is expressed as
	\begin{align}
		\label{runtime2}
		t(\tau\gamma t_\text{CP}+\tau t_\text{DE} + t_\text{EC}), \forall t.
	\end{align}
	The parameters used for these evaluations are listed in Table III. 
	
	In Fig. 6, the mixed setting includes one set of devices with i.i.d. distribution, one set with non-i.i.d. distribution, and a third set where half of the devices have i.i.d. distribution and the other half have non-i.i.d. distribution. The non-i.i.d. distribution in this scenario refers to the case where each device randomly holds data from only two classes, similar to the non-i.i.d. distribution used earlier. In Fig. 7, all devices across the three sets follow this non-i.i.d. distribution, referred to here as the non-i.i.d.1 setting. In Fig. 8, a different non-i.i.d. distribution is used, referred to as the non-i.i.d.2 setting, where each device in any set holds data randomly from only one class. Thus, from Fig. 5 to Fig. 8, the level of data heterogeneity progressively increases. 
	
	As observed, although both algorithms achieve similar performance after convergence in the i.i.d. setting, {\fontfamily{lmtt}\selectfont QHetFed} significantly outperforms {\fontfamily{lmtt}\selectfont Hier-Local-QSGD} in the mixed and non-i.i.d. settings. Additionally, as data heterogeneity increases, the performance gap widens further. The degraded performance of {\fontfamily{lmtt}\selectfont Hier-Local-QSGD} stems from the propagation of errors due to data heterogeneity across multiple local steps in each intra-set iteration, leading local models to converge towards local optima rather than the global optimum. Conversely, {\fontfamily{lmtt}\selectfont
		QHetFed} strategically applies multiple-step local training only at the end of each inter-set iteration, following cloud aggregation. This aggregation involves much more clients than intra-set aggregations, making it potentially more robust against error propagation.
	
	Fig. 9 illustrates the accuracy of {\fontfamily{lmtt}\selectfont
		QHetFed} achieved with the optimized selection of $\tau$ and $\gamma$, as described in Section IV, alongside a random selection of these parameters in the non-i.i.d.1 setting, using the parameters listed in Table III and $q_1 = 11.9$. A delay per iteration $T_\text{di}(\tau, \gamma) \approx 156\ \text{sec}$ is considered, and for the random selection one of the feasible $\tau$ and $\gamma$ pairs are selected. As observed, the optimized parameters lead to notably faster convergence, highlighting the effectiveness of the proposed system optimization.

	\section{Conclusions}
	In this paper, we proposed a new two-level federated learning algorithm tailored to enhance the functionality of hierarchical network structures with multiple sets, employing quantization to facilitate effective communication and specifically addressing the data heterogeneity challenges inherent in IoT systems. This algorithm introduces a novel approach to aggregation, utilizing intra-set gradient and inter-set model parameter aggregation. We provided a comprehensive mathematical methodology for optimality gap analysis of the algorithm, that also incorporates a data heterogeneity metric. Our results demonstrate the negative, but uncorrelated effects of quantization and data heterogeneity. Supported by experimental evidence, our results highlight the enhanced robustness of our hierarchical learning solution compared to the conventional method, with the performance gap widening as the level of heterogeneity increases. Furthermore, for delay-constrained tasks, we derived optimal intra- and inter-set iteration values, demonstrating that these need to be selected by taking the quantization and the communication and computing resources into account.
	
	\appendix
	\section*{Proof of Theorem 1}
	The update of the learning model at the global inter-set iteration $t+1$ is represented as
	\begin{align}
		\label{update}
		&{\mathbf{w}}_{t+1} = \mathbf{w}_t+\frac{1}{N} \sum_{l}^{}  N_l Q_2\biggl( \frac{1}{N_l}\sum_{n}^{} -\mu \sum_{i=0}^{\tau-1}\mathbf{g}_{i,t}^{l}-\nonumber\\&Q_1\biggl(\mu \sum_{j=0}^{\gamma-1}\nabla F_{n}^{l}(\mathbf{w}_{n,\tau,j,t}^{l},\boldsymbol\xi_{n,\tau,j,t}^l)\biggr)\biggr).  
	\end{align}
	From $L$-Lipschitz continuous property in Assumption 2, we have
	\begin{align}
		\label{lip}
		&F(\mathbf{w}_{t+1}) - F(\mathbf{w}_{t}) \leq \nabla F( \mathbf{w}_{t})^\top \left({\mathbf{w}}_{t+1}- {\mathbf{w}}_{t}\right)+\nonumber\\&\frac{L}{2} \|{\mathbf{w}}_{t+1} - {\mathbf{w}}_{t}\|^2.
	\end{align}
	Proceeding by applying the expectation to both sides of \eqref{lip}, we have
	\begin{align}
		\label{lipexp}
		&\mathbb{E}\left\{F({\mathbf{w}}_{t+1})-F({\mathbf{w}}_{t})\right\} \leq \mathbb{E}\left\{\nabla F( \mathbf{w}_{t})^\top \left({\mathbf{w}}_{t+1}- {\mathbf{w}}_{t}\right)\right\}\nonumber\\&+\frac{L}{2} \mathbb{E}\left\{\|{\mathbf{w}}_{t+1} - {\mathbf{w}}_{t}\|^2\right\}.
	\end{align}
	Next, we can expand the first term of the RHS in \eqref{lipexp} as
	\begin{align}
		\label{curlexp}
		&\mathbb{E}\left\{\nabla F( \mathbf{w}_{t})^\top \left({\mathbf{w}}_{t+1}- {\mathbf{w}}_{t}\right)\right\} = \mathbb{E}\biggl\{\nabla F( \mathbf{w}_{t})^\top\frac{1}{N} \sum_{l}^{}  N_l\nonumber\\
		& Q_2\biggl( \frac{1}{N_l}\sum_{n}^{} -\mu \sum_{i=0}^{\tau-1}\mathbf{g}_{i,t}^{l}-Q_1\biggl(\mu_t \sum_{j=0}^{\gamma-1}\nabla F_{n}^{l}(\mathbf{w}_{n,\tau,j,t}^{l},\boldsymbol\xi_{n,\tau,j,t}^l)\nonumber\\&\biggr)\biggr)\biggr\}= -\mu\frac{1}{N} \sum_{l}^{}N_l \sum_{i=0}^{\tau-1}\mathbb{E}\left\{\nabla F( \mathbf{w}_{t})^\top\mathbf{g}_{i,t}^{l}\right\}-\frac{\mu}{N}\times\nonumber\\
		& \sum_{l}^{}\sum_{n}^{}\sum_{j=0}^{\gamma-1}\mathbb{E}\left\{\nabla F( \mathbf{w}_{t})^\top\nabla F_{n}^{l}(\mathbf{w}_{n,\tau,j,t}^{l},\boldsymbol\xi_{n,\tau,j,t}^l)\right\},
	\end{align}
	where
	\begin{align}
		\label{innersum}
		&\mathbb{E}\left\{\nabla F( \mathbf{w}_{t})^\top\mathbf{g}_{i,t}^{l}\right\} = \mathbb{E}\biggl\{\nabla F( \mathbf{w}_{t})^\top\frac{1}{N_l} \sum_{n \in {\cal C}^l}^{} \nonumber\\& Q_1(\nabla F_{n}^{l}(\mathbf{w}_{n,i,t}^l,\boldsymbol\xi_{n,i,t}^l))\biggr\} = \frac{1}{N_l} \sum_{n \in {\cal C}^l}^{} \mathbb{E}\bigl\{\nabla F( \mathbf{w}_{t})^\top\nonumber\\
		&\nabla F_{n}^{l}(\mathbf{w}_{n,i,t}^l,\boldsymbol\xi_{n,i,t}^l)\bigr\}= \frac{1}{N_l} \sum_{n \in {\cal C}^l}^{} \mathbb{E}\left\{\nabla F( \mathbf{w}_{t})^\top\nabla F_n^l(\mathbf{w}_{n,i,t}^l)\right\}.
	\end{align}
	Applying the equality $\|\mathbf{a}_1-\mathbf{a}_2\|^2 = \|\mathbf{a}_1\|^2 + \|\mathbf{a}_2\|^2 - 2\mathbf{a}_1^\top \mathbf{a}_2$ to any vectors $\mathbf{a}_1$ and $\mathbf{a}_2$, we can express the term within the sum \eqref{innersum} as 
	\begin{align}
		\label{diffexp}
		&\mathbb{E}\left\{\nabla F( {\mathbf{w}}_{t})^\top \nabla F_n^l(\mathbf{w}_{n,i,t}^l)\right\} = \frac{1}{2} \mathbb{E}\left\{\|\nabla F( {\mathbf{w}}_{t})\|^2\right\} + \frac{1}{2}\nonumber
		\end{align}
		\begin{align}
		& \mathbb{E}\left\{\|\nabla F_n^l(\mathbf{w}_{n,i,t}^l)\|^2\right\} - \frac{1}{2} \mathbb{E}\left\{\|\nabla F( {\mathbf{w}}_{t})- \nabla F_n^l(\mathbf{w}_{n,i,t}^l)\|^2\right\}.
	\end{align}
	Based on Assumption 2 and the definition of client data heterogeneity, the final term in \eqref{diffexp} is bounded as
	\begin{align}
		\label{diffnorm}
		&\mathbb{E}\left\{\|\nabla F( {\mathbf{w}}_{t})- \nabla F_n^l(\mathbf{w}_{n,i,t}^l)\|^2\right\} = \mathbb{E}\bigl\{\|\nabla F( {\mathbf{w}}_{t})-\nonumber\\
		&\nabla F_n^l( {\mathbf{w}}_{t})+\nabla F_n^l( {\mathbf{w}}_{t})- \nabla F_n^l(\mathbf{w}_{n,i,t}^l)\|^2\bigr\} \leq\nonumber\\
		& G^2 + L^2 \mathbb{E}\left\{\|\mathbf{w}_{t}-\mathbf{w}_{n,i,t}^l\|^2\right\}= G^2 + L^2\times\nonumber\\
		& \mathbb{E}\biggl\{\biggl\Vert-\mu \sum_{j=0}^{i-1} \mathbf{g}_{j,t}^l\biggr \Vert^2\biggr\}= G^2 + L^2 \mu^2 \mathbb{E}\biggl\{\biggl\Vert\sum_{j=0}^{i-1}{\mathbf{g}}_{j,t}^l \biggr\Vert^2\biggr\}.
	\end{align}
	Utilizing the equality $\mathbb{E}\left\{\|\mathbf{a}\|^2\right\} = \|\mathbb{E}\left\{\mathbf{a}\right\}\|^2 + \mathbb{E}\left\{\|\mathbf{a}-\mathbb{E}\left\{\mathbf{a}\right\}\|^2\right\}$ for any vector $\mathbf{a}$, it follows that
	\begin{align}
		\label{normexp}
		&\mathbb{E}\biggl\{\biggl\Vert\sum_{j=0}^{i-1}{\mathbf{g}}_{j,t}^l \biggr\Vert^2\biggr\} =\mathbb{E}\biggl\{\biggl\Vert\sum_{j=0}^{i-1}\frac{1}{N_l} \sum_{n \in {\cal C}^l}^{} \nonumber\\&Q_1(\nabla F_{n}^{l}(\mathbf{w}_{n,j,t}^l,\boldsymbol\xi_{n,j,t}^l)) \biggr\Vert^2\biggr\} = \mathbb{E}\biggl\{\biggl\Vert\sum_{j=0}^{i-1}\frac{1}{N_l} \sum_{n \in {\cal C}^l}^{} \nonumber\\&\nabla F_n^l(\mathbf{w}_{n,j,t}^l) \biggr\Vert^2\biggr\}+\mathbb{E}\biggl\{\biggl\Vert\sum_{j=0}^{i-1}\frac{1}{N_l} \sum_{n \in {\cal C}^l}^{} \Bigl(\nonumber\\&Q_1(\nabla F_{n}^{l}(\mathbf{w}_{n,j,t}^l,\boldsymbol\xi_{n,j,t}^l))-\nabla F_n^l(\mathbf{w}_{n,j,t}^l))\Bigr) \biggr\Vert^2\biggr\},
	\end{align}
	where the first term of RHS can be bounded as
	\begin{align}
		\label{RHS1}
		&\mathbb{E}\biggl\{\biggl\Vert\sum_{j=0}^{i-1}\frac{1}{N_l} \sum_{n \in {\cal C}^l}^{} \nabla F_n^l(\mathbf{w}_{n,j,t}^l) \biggr\Vert^2\biggr\} \stackrel{(a)}{\leq}  i\sum_{j=0}^{i-1}\nonumber\\
		&\mathbb{E}\biggl\{\biggl\Vert\frac{1}{N_l} \sum_{n \in {\cal C}^l}^{} \nabla F_n^l(\mathbf{w}_{n,j,t}^l) \biggr\Vert^2\biggr\}\stackrel{(b)}{\leq}  i \sum_{j=0}^{i-1}\frac{1}{N_l} \sum_{n \in {\cal C}^l}^{} \nonumber\\
		&\mathbb{E}\left\{\left\Vert\nabla F_n^l(\mathbf{w}_{n,j,t}^l) \right\Vert^2\right\},
	\end{align}
	where $(a)$ is derived from the arithmetic-geometric mean inequality, specifically, $(\sum_{i=1}^{I}a_i)^2 \leq I \sum_{i=1}^{I}a_i^2$, and $(b)$ results from the convexity of the $\|.\|^2$ function. The second term of RHS in \eqref{normexp} can be bounded as
	\begin{align}
		\label{RHS2}
		&\mathbb{E}\biggl\{\biggl\Vert\sum_{j=0}^{i-1}\frac{1}{N_l} \sum_{n \in {\cal C}^l}^{} \left(Q_1(\nabla F_{n}^{l}(\mathbf{w}_{n,j,t}^l,\boldsymbol\xi_{n,j,t}^l))-\nabla F_n^l(\mathbf{w}_{n,j,t}^l))\right)\nonumber\\& \biggr\Vert^2\biggr\}\stackrel{(c)}{=} \sum_{j=0}^{i-1}\frac{1}{N_l^2} \sum_{n \in {\cal C}^l}^{} \mathbb{E}\bigl\{\bigl\Vert\ Q_1(\nabla F_{n}^{l}(\mathbf{w}_{n,j,t}^l,\boldsymbol\xi_{n,j,t}^l))-\nonumber\\&\nabla F_n^l(\mathbf{w}_{n,j,t}^l,\boldsymbol\xi_{n,j,t}^l) \bigr\Vert^2\bigr\}+\sum_{j=0}^{i-1}\frac{1}{N_l^2} \sum_{n \in {\cal C}^l}^{} \nonumber\\
		&\mathbb{E}\bigl\{\bigl\Vert \nabla F_{n}^{l}(\mathbf{w}_{n,j,t}^l,\boldsymbol\xi_{n,j,t}^l)-\nabla F_n^l(\mathbf{w}_{n,j,t}^l) \bigr\Vert^2\bigr\}\stackrel{(d)}{\leq} \frac{q_1}{N_l^2}\times\nonumber\\
		&\sum_{j=0}^{i-1} \sum_{n \in {\cal C}^l}^{} \mathbb{E}\bigl\{\bigl\Vert\ \nabla F_{n}^{l}(\mathbf{w}_{n,j,t}^l,\boldsymbol\xi_{n,j,t}^l) \bigr\Vert^2\bigr\}+\frac{\sigma^2}{B}\frac{i}{N_l}
		,
	\end{align}
	where
	\begin{align}
		&\mathbb{E}\left\{\left\Vert\ \nabla F_{n}^{l}(\mathbf{w}_{n,j,t}^l,\boldsymbol\xi_{n,j,t}^l) \right\Vert^2\right\} = \mathbb{E}\left\{\left\Vert\ \nabla F_n^l(\mathbf{w}_{n,j,t}^l) \right\Vert^2\right\}+\nonumber
						\end{align}
		\begin{align}
		&\mathbb{E}\left\{\left\Vert\ \nabla F_{n}^{l}(\mathbf{w}_{n,j,t}^l,\boldsymbol\xi_{n,j,t}^l)-\nabla F_n^l(\mathbf{w}_{n,j,t}^l) \right\Vert^2\right\} \leq \nonumber\\
		&\mathbb{E}\left\{\left\Vert\ \nabla F_n^l(\mathbf{w}_{n,j,t}^l) \right\Vert^2\right\}+\frac{\sigma^2}{B}
		.
	\end{align}
	The step $(c)$ comes from the independence conditioned on batches $\boldsymbol\xi_{n,j,t}^l$
	for any two distinct values of $n$, $l$, $j$, or $t$. Then, $(d)$ comes from the Assumptions 1 and 3. Replacing \eqref{RHS1} and \eqref{RHS2} in \eqref{normexp} and then replacing the result in \eqref{diffnorm}, we have
	\begin{align}
		\label{normdiff}
		&\mathbb{E}\left\{\|\nabla F( {\mathbf{w}}_{t})- \nabla F(\mathbf{w}_{n,i,t}^l)\|^2\right\} \leq G^2 + L^2 \mu^2 \left(\frac{i}{N_l}+\frac{q_1}{N_l^2}\right)\nonumber\\& \sum_{j=0}^{i-1} \sum_{n \in {\cal C}^l}^{} \mathbb{E}\left\{\left\Vert\nabla F_n^l(\mathbf{w}_{n,j,t}^l) \right\Vert^2\right\}+L^2 \mu^2 (1+q_1)\frac{\sigma^2}{B}\frac{i}{N_l},
	\end{align}
	and then replacing \eqref{normdiff} in \eqref{diffexp} and replacing the result in \eqref{innersum}, we obtain the following bound  
	\begin{align}
		\label{1RHS1}
		&-\mu\frac{1}{N} \sum_{l}^{}N_l \sum_{i=0}^{\tau-1}\mathbb{E}\left\{\nabla F( \mathbf{w}_{t})^\top\mathbf{g}_{i,t}^{l}\right\} \leq -\frac{\mu \tau}{2} \times\nonumber\\&\mathbb{E}\left\{\|\nabla F( {\mathbf{w}}_{t})\|^2\right\}-\frac{\mu}{2N} \sum_{l}^{}\sum_{i=0}^{\tau-1}\sum_{n}^{}\mathbb{E}\left\{\|\nabla F_n^l(\mathbf{w}_{n,i,t}^l)\|^2\right\}\nonumber\\&+\frac{\mu \tau}{2}G^2+\frac{L^2\mu^3}{2N}\sum_{l} \sum_{i=0}^{\tau-1} \left({i+\frac{q_1}{N_l}}\right)\sum_{j=0}^{i-1} \sum_{n} \nonumber\\&\mathbb{E}\left\{\|\nabla F_n^l(\mathbf{w}_{n,j,t}^l)\|^2\right\} + \frac{L^2 \mu^3}{2N}C(1+q_1)\frac{\sigma^2}{B} \frac{\tau(\tau-1)}{2}.
	\end{align}
	Next, we can bound the second term in RHS of \eqref{curlexp} as follows.
	\begin{align}
		\label{new_prod}
		&-\mu\frac{1}{N} \sum_{l}^{}\sum_{n}^{}\sum_{j=0}^{\gamma-1}\mathbb{E}\left\{\nabla F( \mathbf{w}_{t})^\top\nabla F_{n}^{l}(\mathbf{w}_{n,\tau,j,t}^{l},\boldsymbol\xi_{n,\tau,j,t}^l)\right\} \nonumber\\
		&= -\mu\frac{1}{N} \sum_{l}^{}\sum_{n}^{}\sum_{j=0}^{\gamma-1}\mathbb{E}\left\{\nabla F( \mathbf{w}_{t})^\top\nabla F_n^l(\mathbf{w}_{n,\tau,j,t}^{l})\right\},
	\end{align}
	where
	\begin{align}
		&\mathbb{E}\left\{\nabla F( \mathbf{w}_{t})^\top\nabla F_n^l(\mathbf{w}_{n,\tau,j,t}^{l})\right\} = \frac{1}{2} \mathbb{E}\left\{\|\nabla F( {\mathbf{w}}_{t})\|^2\right\} +\nonumber\\& \frac{1}{2} \mathbb{E}\left\{\|\nabla F_n^l(\mathbf{w}_{n,\tau,j,t}^l)\|^2\right\} - \frac{1}{2} \times \nonumber\\&\mathbb{E}\left\{\|\nabla F( {\mathbf{w}}_{t})- \nabla F_n^l(\mathbf{w}_{n,\tau,j,t}^l)\|^2\right\},
	\end{align}
	where from Definition 1
	\begin{align}
		&\mathbb{E}\left\{\|\nabla F( {\mathbf{w}}_{t})- \nabla F_n^l(\mathbf{w}_{n,\tau,j,t}^l)\|^2\right\} \leq G^2 +\nonumber\\& L^2 \mu^2 \mathbb{E}\biggl\{\biggl\Vert\sum_{i=0}^{\tau-1}{\mathbf{g}}_{i,t}^l +\sum_{p=0}^{j-1} \nabla F_{n}^{l}(\mathbf{w}_{n,\tau,p,t}^{l},\boldsymbol\xi_{n,\tau,p,t}^l) \biggr\Vert^2\biggr\},
	\end{align}
	where
	\begin{align}
		&\mathbb{E}\biggl\{\biggl\Vert\sum_{i=0}^{\tau-1}{\mathbf{g}}_{i,t}^l +\sum_{p=0}^{j-1} \nabla F_{n}^{l}(\mathbf{w}_{n,\tau,p,t}^{l},\boldsymbol\xi_{n,\tau,p,t}^l) \biggr\Vert^2\biggr\} =\nonumber\\& \mathbb{E}\biggl\{\biggl\Vert\sum_{i=0}^{\tau-1}\frac{1}{N_l} \sum_{n \in {\cal C}^l}^{} Q_1(\nabla F_{n}^{l}(\mathbf{w}_{n,i,t}^l,\boldsymbol\xi_{n,i,t}^l)) +\sum_{p=0}^{j-1} \nonumber
						\end{align}
		\begin{align}
		&\nabla F_{n}^{l}(\mathbf{w}_{n,\tau,p,t}^{l},\boldsymbol\xi_{n,\tau,p,t}^l) \biggr\Vert^2\biggr\} = \mathbb{E}\biggl\{\biggl\Vert\sum_{i=0}^{\tau-1}\frac{1}{N_l} \sum_{n \in {\cal C}^l}^{} \nabla F_n^l(\mathbf{w}_{n,i,t}^l) \nonumber\\
		&+\sum_{p=0}^{j-1} \nabla F_n^l(\mathbf{w}_{n,\tau,p,t}^{l}) \biggr\Vert^2\biggr\}+\frac{1}{N_l^2}\sum_{i=0}^{\tau-1} \sum_{n \in {\cal C}^l}^{}\nonumber\\
		& \mathbb{E}\left\{\left\Vert Q_1(\nabla F_{n}^{l}(\mathbf{w}_{n,i,t}^l,\boldsymbol\xi_{n,i,t}^l)) -\nabla F_n^l(\mathbf{w}_{n,i,t}^l)\right\Vert^2\right\}  +\nonumber\\
		&\sum_{p=0}^{j-1} \mathbb{E}\left\{\left\Vert\nabla F_{n}^{l}(\mathbf{w}_{n,\tau,p,t}^{l},\boldsymbol\xi_{n,\tau,p,t}^l) -\nabla F_n^l(\mathbf{w}_{n,\tau,p,t}^{l})\right\Vert^2\right\},
	\end{align}
	where from the arithmetic-geometric mean inequality
	\begin{align}
		&\mathbb{E}\biggl\{\biggl\Vert\sum_{i=0}^{\tau-1}\frac{1}{N_l} \sum_{n \in {\cal C}^l}^{} \nabla F_n^l(\mathbf{w}_{n,i,t}^l) +\sum_{p=0}^{j-1} \nabla F_n^l(\mathbf{w}_{n,\tau,p,t}^{l}) \biggr\Vert^2\biggr\}  \nonumber\\
		&\leq\tau \sum_{i=0}^{\tau-1}\mathbb{E}\biggl\{\biggl\Vert\frac{1}{N_l} \sum_{n \in {\cal C}^l}^{} \nabla F_n^l(\mathbf{w}_{n,i,t}^l)\biggr\Vert^2\biggr\} +j\sum_{p=0}^{j-1} \nonumber\\
		&\mathbb{E}\left\{\left\Vert\nabla F_n^l(\mathbf{w}_{n,\tau,p,t}^{l})\right\Vert^2\right\} \leq \tau \sum_{i=0}^{\tau-1}\frac{1}{N_l} \sum_{n \in {\cal C}^l}^{} \nonumber\\&\mathbb{E}\Bigl\{\left\Vert\nabla F_n^l(\mathbf{w}_{n,i,t}^l)\right\Vert^2\Bigr\} +j\sum_{p=0}^{j-1} \mathbb{E}\left\{\left\Vert\nabla F_n^l(\mathbf{w}_{n,\tau,p,t}^{l})\right\Vert^2\right\},
	\end{align}
	and
	\begin{align}
		&\frac{1}{N_l^2}\sum_{i=0}^{\tau-1} \sum_{n \in {\cal C}^l}^{} \mathbb{E}\bigl\{\bigl\Vert Q_1(\nabla F_{n}^{l}(\mathbf{w}_{n,i,t}^l,\boldsymbol\xi_{n,i,t}^l)) -\nabla F_n^l(\mathbf{w}_{n,i,t}^l)\bigr\Vert^2\nonumber\\
		&\bigr\} =  \frac{1}{N_l^2}\sum_{i=0}^{\tau-1} \sum_{n \in {\cal C}^l}^{} \mathbb{E}\bigl\{\bigl\Vert Q_1(\nabla F_{n}^{l}(\mathbf{w}_{n,i,t}^l,\boldsymbol\xi_{n,i,t}^l)) -\nonumber\\
		&\nabla F_n^l(\mathbf{w}_{n,i,t}^l,\boldsymbol\xi_{n,i,t}^l)\bigr\Vert^2\bigr\}+\frac{1}{N_l^2}\sum_{i=0}^{\tau-1} \sum_{n \in {\cal C}^l}^{} \nonumber\\
		&\mathbb{E}\left\{\left\Vert \nabla F_{n}^{l}(\mathbf{w}_{n,i,t}^l,\boldsymbol\xi_{n,i,t}^l) -\nabla F_n^l(\mathbf{w}_{n,i,t}^l)\right\Vert^2\right\}=\frac{q_1}{N_l^2}\sum_{i=0}^{\tau-1} \sum_{n \in {\cal C}^l}^{} \nonumber\\
		&\mathbb{E}\left\{\left\Vert \nabla F_{n}^{l}(\mathbf{w}_{n,i,t}^l,\boldsymbol\xi_{n,i,t}^l)\right\Vert^2\right\}+\frac{\sigma^2}{B} \frac{\tau}{N_l} =\frac{q_1}{N_l^2}\sum_{i=0}^{\tau-1} \sum_{n \in {\cal C}^l}^{}\nonumber\\
		& \mathbb{E}\left\{\left\Vert \nabla F_n^l(\mathbf{w}_{n,i,t}^l)\right\Vert^2\right\}+\frac{\sigma^2}{B} \frac{\tau}{N_l}(1+q_1).
	\end{align}
	Thus, we obtain
	\begin{align}
		\label{new_prod}
		&-\mu\frac{1}{N} \sum_{l}^{}\sum_{n}^{}\sum_{j=0}^{\gamma-1}\mathbb{E}\left\{\nabla F( \mathbf{w}_{t})^\top\nabla F_{n}^{l}(\mathbf{w}_{n,\tau,j,t}^{l},\boldsymbol\xi_{n,\tau,j,t}^l)\right\} =\nonumber\\
		& -\frac{\mu\gamma}{2}  \mathbb{E}\left\{\|\nabla F( {\mathbf{w}}_{t})\|^2\right\}  -\mu\frac{1}{2N} \sum_{l}^{}\sum_{n}^{}\sum_{j=0}^{\gamma-1}\nonumber\\
		& \mathbb{E}\left\{\|\nabla F_n^l(\mathbf{w}_{n,\tau,j,t}^l)\|^2\right\} +\frac{L^2\mu^3}{2N} \tau \gamma \sum_{l}^{}   \sum_{i=0}^{\tau-1} \sum_{n \in {\cal C}^l}^{}\nonumber\\
		& \mathbb{E}\left\{\left\Vert\nabla F_n^l(\mathbf{w}_{n,i,t}^l)\right\Vert^2\right\} +\frac{L^2\mu^3}{2N} \sum_{l}^{}\sum_{n}^{}\sum_{j=0}^{\gamma-1}  j\sum_{p=0}^{j-1} \nonumber
				\end{align}
		\begin{align}
		&\mathbb{E}\left\{\left\Vert\nabla F_n^l(\mathbf{w}_{n,\tau,p,t}^{l})\right\Vert^2\right\}+ \frac{L^2\mu^3}{2N} q_1\gamma \sum_{l}^{} \frac{1}{N_l}\sum_{i=0}^{\tau-1} \sum_{n \in {\cal C}^l}^{} \nonumber\\&\mathbb{E}\left\{\left\Vert \nabla F_n^l(\mathbf{w}_{n,i,t}^l)\right\Vert^2\right\}+\frac{L^2\mu^3}{2N} \tau \gamma C\frac{\sigma^2}{B}(1+q_1)+\nonumber\\&\frac{L^2\mu^3}{2} \frac{\sigma^2}{B} \frac{\gamma(\gamma-1)}{2}+\frac{\mu\gamma}{2}G^2.
	\end{align}
	Next, we bound the second term of the RHS in \eqref{lipexp} as 
	\begin{align}
		\label{normtwo}
		&\mathbb{E}\left\{\|{\mathbf{w}}_{t+1} - {\mathbf{w}}_{t}\|^2\right\} = \mathbb{E}\biggl\{\biggl\Vert\frac{1}{N} \sum_{l}^{}  N_l Q_2\biggl( \frac{1}{N_l}\sum_{n}^{} -\mu \sum_{i=0}^{\tau-1}\nonumber\\
		&\mathbf{g}_{i,t}^{l}-Q_1\biggl(\mu \sum_{j=0}^{\gamma-1}\nabla F_{n}^{l}(\mathbf{w}_{n,\tau,j,t}^{l},\boldsymbol\xi_{n,\tau,j,t}^l)\biggr)\biggr)\biggr\Vert^2\biggr\}= \mathbb{E}\biggl\{\biggl\Vert\frac{1}{N} \nonumber\\&\sum_{l}^{}  \sum_{n}^{} -\mu \sum_{i=0}^{\tau-1}\mathbf{g}_{i,t}^{l}-Q_1\biggl(\mu \sum_{j=0}^{\gamma-1}\nabla F_{n}^{l}(\mathbf{w}_{n,\tau,j,t}^{l},\boldsymbol\xi_{n,\tau,j,t}^l)\biggr)\biggr\Vert^2\nonumber\\&\biggr\}+\frac{q_2}{N^2} \sum_{l}^{}   \mathbb{E}\biggl\{\biggl\Vert  \sum_{n}^{} -\mu \sum_{i=0}^{\tau-1}\mathbf{g}_{i,t}^{l}-Q_1\biggl(\mu \sum_{j=0}^{\gamma-1}\nonumber\\
		&\nabla F_{n}^{l}(\mathbf{w}_{n,\tau,j,t}^{l},\boldsymbol\xi_{n,\tau,j,t}^l)\biggr)\biggr\Vert^2\biggr\}=\mathbb{E}\biggl\{\biggl\Vert-\frac{\mu}{N} \sum_{l}^{}  \sum_{n}^{}  \sum_{i=0}^{\tau-1}\mathbf{g}_{i,t}^{l}\nonumber\\&-\frac{\mu}{N} \sum_{l}^{}  \sum_{n}^{} \sum_{j=0}^{\gamma-1}\nabla F_{n}^{l}(\mathbf{w}_{n,\tau,j,t}^{l},\boldsymbol\xi_{n,\tau,j,t}^l)\biggr\Vert^2\biggr\}\nonumber\\
		&+\frac{q_1}{N^2} \sum_{l}^{}  \sum_{n}^{} \mathbb{E}\biggl\{\biggl\Vert-\mu \sum_{j=0}^{\gamma-1}\nabla F_{n}^{l}(\mathbf{w}_{n,\tau,j,t}^{l},\boldsymbol\xi_{n,\tau,j,t}^l)\biggr\Vert^2\biggr\}\nonumber\\
		&+\frac{q_2}{N^2} \sum_{l}^{}   \mathbb{E}\biggl\{\biggl\Vert  -\mu\sum_{n}^{}  \sum_{i=0}^{\tau-1}\mathbf{g}_{i,t}^{l}-\mu \sum_{n}^{} \sum_{j=0}^{\gamma-1}\nonumber\\
		&\nabla F_{n}^{l}(\mathbf{w}_{n,\tau,j,t}^{l},\boldsymbol\xi_{n,\tau,j,t}^l)\biggr\Vert^2\biggr\}+\frac{q_2q_1}{N^2} \sum_{l}^{}     \sum_{n}^{}\mathbb{E}\biggl\{\biggl\Vert-\mu \sum_{j=0}^{\gamma-1}\nonumber\\
		&\nabla F_{n}^{l}(\mathbf{w}_{n,\tau,j,t}^{l},\boldsymbol\xi_{n,\tau,j,t}^l)\biggr\Vert^2\biggr\}=\mathbb{E}\biggl\{\biggl\Vert-\frac{\mu}{N} \sum_{l}^{}  \sum_{n}^{}  \sum_{i=0}^{\tau-1}\mathbf{g}_{i,t}^{l}-\nonumber\\
		&\frac{\mu}{N} \sum_{l}^{}  \sum_{n}^{} \sum_{j=0}^{\gamma-1}\nabla F_{n}^{l}(\mathbf{w}_{n,\tau,j,t}^{l},\boldsymbol\xi_{n,\tau,j,t}^l)\biggr\Vert^2\biggr\}+\frac{q_2}{N^2} \sum_{l}^{} \mathbb{E}\biggl\{\biggl\Vert\nonumber\\
		&  -\mu\sum_{n}^{}  \sum_{i=0}^{\tau-1}\mathbf{g}_{i,t}^{l}-\mu \sum_{n}^{} \sum_{j=0}^{\gamma-1}\nabla F_{n}^{l}(\mathbf{w}_{n,\tau,j,t}^{l},\boldsymbol\xi_{n,\tau,j,t}^l)\biggr\Vert^2\biggr\}+\nonumber\\
		&\frac{(1+q_2)q_1}{N^2} \sum_{l}^{}     \sum_{n}^{}\mathbb{E}\biggl\{\biggl\Vert-\mu \sum_{j=0}^{\gamma-1}\nabla F_{n}^{l}(\mathbf{w}_{n,\tau,j,t}^{l},\boldsymbol\xi_{n,\tau,j,t}^l)\biggr\Vert^2\biggr\},
	\end{align}
	where
	\begin{align}
		&\mathbb{E}\biggl\{\biggl\Vert-\frac{\mu}{N} \sum_{l}^{}  \sum_{n}^{}  \sum_{i=0}^{\tau-1}\mathbf{g}_{i,t}^{l}-\frac{\mu}{N} \sum_{l}^{}  \sum_{n}^{} \sum_{j=0}^{\gamma-1}\nonumber\\
		&\nabla F_{n}^{l}(\mathbf{w}_{n,\tau,j,t}^{l},\boldsymbol\xi_{n,\tau,j,t}^l)\biggr\Vert^2\biggr\}=\frac{\mu^2}{N^2}\mathbb{E}\biggl\{\biggl\Vert \sum_{l}^{}  \sum_{n}^{}  \sum_{i=0}^{\tau-1}\mathbf{g}_{i,t}^{l}\biggr\Vert^2\biggr\}\nonumber\\
		&+\frac{\mu^2}{N^2}\mathbb{E}\biggl\{\biggl\Vert \sum_{l}^{}  \sum_{n}^{} \sum_{j=0}^{\gamma-1}\nabla F_{n}^{l}(\mathbf{w}_{n,\tau,j,t}^{l},\boldsymbol\xi_{n,\tau,j,t}^l)\biggr\Vert^2\biggr\}=\frac{\mu^2}{N^2}\nonumber
						\end{align}
		\begin{align}
		&\mathbb{E}\biggl\{\biggl\Vert \sum_{l}^{}  \sum_{i=0}^{\tau-1} \sum_{n \in {\cal C}^l}^{} Q_1(\nabla F_{n}^{l}(\mathbf{w}_{n,i,t}^l,\boldsymbol\xi_{n,i,t}^l))\biggr\Vert^2\biggr\}+\frac{\mu^2}{N^2}\nonumber\\
		&\mathbb{E}\biggl\{\biggl\Vert \sum_{l}^{}  \sum_{n}^{} \sum_{j=0}^{\gamma-1}\nabla F_{n}^{l}(\mathbf{w}_{n,\tau,j,t}^{l},\boldsymbol\xi_{n,\tau,j,t}^l)\biggr\Vert^2\biggr\}=\frac{\mu^2}{N^2}\nonumber\\
		&\mathbb{E}\biggl\{\biggl\Vert \sum_{l}^{}  \sum_{i=0}^{\tau-1} \sum_{n \in {\cal C}^l}^{} \nabla F_n^l(\mathbf{w}_{n,i,t}^l)\biggr\Vert^2\biggr\}+\frac{\mu^2}{N^2} \sum_{l}^{}  \sum_{i=0}^{\tau-1} \sum_{n \in {\cal C}^l}^{}\nonumber\\
		& \mathbb{E}\biggl\{\biggl\Vert Q_1(\nabla F_{n}^{l}(\mathbf{w}_{n,i,t}^l,\boldsymbol\xi_{n,i,t}^l))-\nabla F_{n}^{l}(\mathbf{w}_{n,i,t}^l,\boldsymbol\xi_{n,i,t}^l)\biggr\Vert^2\biggr\} +\nonumber\\
		&\frac{\mu^2}{N^2} \sum_{l}^{}  \sum_{i=0}^{\tau-1} \sum_{n \in {\cal C}^l}^{} \mathbb{E}\biggl\{\biggl\Vert  F_{n}^{l}(\mathbf{w}_{n,i,t}^l,\boldsymbol\xi_{n,i,t}^l)-\nabla F_n^l(\mathbf{w}_{n,i,t}^l)\biggr\Vert^2\biggr\}\nonumber\\
		&+\frac{\mu^2}{N^2}\mathbb{E}\biggl\{\biggl\Vert \sum_{l}^{}  \sum_{n}^{} \sum_{j=0}^{\gamma-1}\nabla F_n^l(\mathbf{w}_{n,\tau,j,t}^{l})\biggr\Vert^2\biggr\}+\frac{\mu^2}{N^2} \sum_{l}^{}  \sum_{n}^{} \nonumber\\
		&\sum_{j=0}^{\gamma-1}\mathbb{E}\biggl\{\biggl\Vert\nabla F_{n}^{l}(\mathbf{w}_{n,\tau,j,t}^{l},\boldsymbol\xi_{n,\tau,j,t}^l)-\nabla F_n^l(\mathbf{w}_{n,\tau,j,t}^{l})\biggr\Vert^2\biggr\}=\frac{\mu^2}{N} \nonumber\\
		&\tau\sum_{l}^{}  \sum_{i=0}^{\tau-1} \sum_{n \in {\cal C}^l}^{} \mathbb{E}\biggl\{\biggl\Vert\nabla F_n^l(\mathbf{w}_{n,i,t}^l)\biggr\Vert^2\biggr\}+\frac{\mu^2}{N^2}q_1 \sum_{l}^{}  \sum_{i=0}^{\tau-1} \sum_{n \in {\cal C}^l}^{} \nonumber\\
		&\mathbb{E}\left\{\left\Vert \nabla F_n^l(\mathbf{w}_{n,i,t}^l)\right\Vert^2\right\}+\frac{\mu^2}{N}q_1\tau\frac{\sigma^2}{B} +\frac{\mu^2}{N}\frac{\sigma^2}{B}\tau +\frac{\mu^2}{N}\gamma \sum_{l}^{} \nonumber\\
		& \sum_{n}^{} \sum_{j=0}^{\gamma-1}\mathbb{E}\left\{\left\Vert\nabla F_n^l(\mathbf{w}_{n,\tau,j,t}^{l})\right\Vert^2\right\}+\frac{\mu^2}{N} \frac{\sigma^2}{B}\gamma = \frac{\mu^2}{N}(\tau+\frac{q_1}{N})\nonumber\\
		& \sum_{l}^{}  \sum_{i=0}^{\tau-1} \sum_{n \in {\cal C}^l}^{} \mathbb{E}\left\{\left\Vert\nabla F_n^l(\mathbf{w}_{n,i,t}^l)\right\Vert^2\right\}+\frac{\mu^2}{N}\gamma \sum_{l}^{}  \sum_{n}^{} \sum_{j=0}^{\gamma-1}\nonumber\\
		&\mathbb{E}\left\{\left\Vert\nabla F_n^l(\mathbf{w}_{n,\tau,j,t}^{l})\right\Vert^2\right\}+\frac{\mu^2}{N} \frac{\sigma^2}{B}(\tau+q_1\tau+\gamma),
	\end{align}
	and
	\begin{align}
		&\frac{q_2}{N^2}\mu^2 \sum_{l}^{}   \mathbb{E}\biggl\{\biggl\Vert  \sum_{n}^{}  \sum_{i=0}^{\tau-1}\mathbf{g}_{i,t}^{l}+ \sum_{n}^{} \sum_{j=0}^{\gamma-1}\nabla F_{n}^{l}(\mathbf{w}_{n,\tau,j,t}^{l},\boldsymbol\xi_{n,\tau,j,t}^l)\nonumber\\&\biggr\Vert^2\biggr\}=\frac{q_2}{N^2}\mu^2 \sum_{l}^{}   \mathbb{E}\biggl\{\biggl\Vert   \sum_{i=0}^{\tau-1}\sum_{n \in {\cal C}^l}^{} Q_1(\nabla F_{n}^{l}(\mathbf{w}_{n,i,t}^l,\boldsymbol\xi_{n,i,t}^l))\biggr\Vert^2\biggr\}\nonumber\\
		&+ \frac{q_2}{N^2}\mu^2 \sum_{l}^{}   \mathbb{E}\biggl\{\biggl\Vert\sum_{n}^{} \sum_{j=0}^{\gamma-1}\nabla F_{n}^{l}(\mathbf{w}_{n,\tau,j,t}^{l},\boldsymbol\xi_{n,\tau,j,t}^l)\biggr\Vert^2\biggr\}=\frac{q_2}{N^2}\nonumber\\
		&\mu^2 \sum_{l}^{}   \mathbb{E}\biggl\{\biggl\Vert   \sum_{i=0}^{\tau-1}\sum_{n \in {\cal C}^l}^{}  \nabla F_n^l(\mathbf{w}_{n,i,t}^l)\biggr\Vert^2\biggr\}+\frac{q_2}{N^2}\mu^2 \sum_{l}^{}      \sum_{i=0}^{\tau-1}\sum_{n \in {\cal C}^l}^{}\nonumber\\
		& \mathbb{E}\left\{\left\Vert Q_1(\nabla F_{n}^{l}(\mathbf{w}_{n,i,t}^l,\boldsymbol\xi_{n,i,t}^l))-\nabla F_{n}^{l}(\mathbf{w}_{n,i,t}^l,\boldsymbol\xi_{n,i,t}^l)\right\Vert^2\right\}+\nonumber\\
		&\frac{q_2}{N^2}\mu^2 \sum_{l}^{}      \sum_{i=0}^{\tau-1}\sum_{n \in {\cal C}^l}^{} \mathbb{E}\bigl\{\bigl\Vert \nabla F_{n}^{l}(\mathbf{w}_{n,i,t}^l,\boldsymbol\xi_{n,i,t}^l)-\nabla F_n^l(\mathbf{w}_{n,i,t}^l)\nonumber\\
		&\bigr\Vert^2\bigr\}+\frac{q_2}{N^2}\mu^2 \sum_{l}^{}   \mathbb{E}\biggl\{\biggl\Vert\sum_{n}^{} \sum_{j=0}^{\gamma-1}\nabla F_n^l(\mathbf{w}_{n,\tau,j,t}^{l})\biggr\Vert^2\biggr\}+\frac{q_2}{N^2}\mu^2\nonumber\\
		& \sum_{l}^{}   \sum_{n}^{} \sum_{j=0}^{\gamma-1}\mathbb{E}\bigl\{\bigl\Vert\nabla F_{n}^{l}(\mathbf{w}_{n,\tau,j,t}^{l},\boldsymbol\xi_{n,\tau,j,t}^l)-\nabla F_n^l(\mathbf{w}_{n,\tau,j,t}^{l})\bigr\Vert^2\nonumber
						\end{align}
		\begin{align}
		&\bigr\}=\frac{q_2}{N^2}\mu^2 \tau \sum_{l}^{}  N_l  \sum_{i=0}^{\tau-1}\sum_{n \in {\cal C}^l}^{}  \mathbb{E}\left\{\left\Vert \nabla F_n^l(\mathbf{w}_{n,i,t}^l)\right\Vert^2\right\}+\frac{q_2 q_1\mu^2}{N^2}\nonumber\\
		& \sum_{l}^{}  \sum_{i=0}^{\tau-1}\sum_{n \in {\cal C}^l}^{} \mathbb{E}\left\{\left\Vert \nabla F_n^l(\mathbf{w}_{n,i,t}^l)\right\Vert^2\right\} +\frac{q_2 q_1}{N}\tau \mu^2 \frac{\sigma^2}{B}+\frac{q_2}{N}\tau\mu^2\nonumber\\
		& \frac{\sigma^2}{B} +\frac{q_2}{N^2}\gamma \mu^2 \sum_{l}^{}   N_l\sum_{n}^{} \sum_{j=0}^{\gamma-1}\mathbb{E}\left\{\left\Vert\nabla F_n^l(\mathbf{w}_{n,\tau,j,t}^{l})\right\Vert^2\right\}+\frac{q_2}{N}\nonumber\\
		&\gamma \mu^2 \frac{\sigma^2}{B}  = \frac{q_2}{N^2}\mu^2 \tau \sum_{l}^{}  N_l  \sum_{i=0}^{\tau-1}\sum_{n \in {\cal C}^l}^{}  \mathbb{E}\left\{\left\Vert \nabla F_n^l(\mathbf{w}_{n,i,t}^l)\right\Vert^2\right\}+\nonumber\\
		&\frac{q_2 q_1}{N^2}\mu^2 \sum_{l}^{}  \sum_{i=0}^{\tau-1}\sum_{n \in {\cal C}^l}^{} \mathbb{E}\left\{\left\Vert \nabla F_n^l(\mathbf{w}_{n,i,t}^l)\right\Vert^2\right\}+\frac{q_2}{N^2}\gamma \mu^2 \sum_{l}^{}   N_l\nonumber\\&\sum_{n}^{} \sum_{j=0}^{\gamma-1}\mathbb{E}\left\{\left\Vert\nabla F_n^l(\mathbf{w}_{n,\tau,j,t}^{l})\right\Vert^2\right\} + \frac{q_2}{N}(\tau+q_1\tau+\gamma) \mu^2 \frac{\sigma^2}{B},
	\end{align}
	and
	\begin{align}
		&\frac{(1+q_2)q_1}{N^2}\mu^2 \sum_{l}^{}     \sum_{n}^{}\mathbb{E}\biggl\{\biggl\Vert \sum_{j=0}^{\gamma-1}\nabla F_{n}^{l}(\mathbf{w}_{n,\tau,j,t}^{l},\boldsymbol\xi_{n,\tau,j,t}^l)\biggr\Vert^2\biggr\} \nonumber\\
		& = \frac{(1+q_2)q_1}{N^2}\mu^2 \sum_{l}^{}     \sum_{n}^{}\mathbb{E}\biggl\{\biggl\Vert \sum_{j=0}^{\gamma-1}\nabla F_n^l(\mathbf{w}_{n,\tau,j,t}^{l})\biggr\Vert^2\biggr\}\nonumber\\
		&+\frac{(1+q_2)q_1}{N^2}\mu^2 \sum_{l}^{}     \sum_{n}^{} \sum_{j=0}^{\gamma-1}\mathbb{E}\bigl\{\bigl\Vert\nabla F_n^l(\mathbf{w}_{n,\tau,j,t}^{l},\boldsymbol\xi_{n,\tau,j,t}^{l}) - \nonumber\\
		&\nabla F_n^l(\mathbf{w}_{n,\tau,j,t}^{l})\bigr\Vert^2\bigr\} =   \frac{(1+q_2)q_1}{N^2}\gamma \mu^2 \sum_{l}^{}     \sum_{n}^{} \sum_{j=0}^{\gamma-1}\nonumber\\
		&\mathbb{E}\left\{\left\Vert\nabla F_n^l(\mathbf{w}_{n,\tau,j,t}^{l})\right\Vert^2\right\}+\frac{(1+q_2)q_1}{N}\gamma \mu^2\frac{\sigma^2}{B}.
	\end{align}
	
	Thus, we obtain \eqref{normtwo} as
	\begin{align}
		\label{normtwofinal}
		&\mathbb{E}\left\{\|{\mathbf{w}}_{t+1} - {\mathbf{w}}_{t}\|^2\right\} =\frac{\mu^2}{N}(\tau+\frac{q_1}{N}) \sum_{l}^{}  \sum_{i=0}^{\tau-1} \sum_{n \in {\cal C}^l}^{}\nonumber\\
		& \mathbb{E}\left\{\left\Vert\nabla F_n^l(\mathbf{w}_{n,i,t}^l)\right\Vert^2\right\}+\frac{\mu^2}{N}\gamma \sum_{l}^{}  \sum_{n}^{} \sum_{j=0}^{\gamma-1}\nonumber\\
		&\mathbb{E}\left\{\left\Vert\nabla F_n^l(\mathbf{w}_{n,\tau,j,t}^{l})\right\Vert^2\right\}+\frac{\mu^2}{N} \frac{\sigma^2}{B}(\tau+q_1\tau+\gamma)+\frac{q_2}{N^2}\mu^2 \tau\nonumber\\
		& \sum_{l}^{}  N_l  \sum_{i=0}^{\tau-1}\sum_{n \in {\cal C}^l}^{}  \mathbb{E}\left\{\left\Vert \nabla F_n^l(\mathbf{w}_{n,i,t}^l)\right\Vert^2\right\}+\frac{q_2 q_1}{N^2}\mu^2 \sum_{l}^{}  \sum_{i=0}^{\tau-1}\nonumber\\&\sum_{n \in {\cal C}^l}^{} \mathbb{E}\left\{\left\Vert \nabla F_n^l(\mathbf{w}_{n,i,t}^l)\right\Vert^2\right\}+\frac{q_2}{N^2}\gamma \mu^2 \sum_{l}^{}   N_l\sum_{n}^{} \sum_{j=0}^{\gamma-1}\nonumber\\
		&\mathbb{E}\left\{\left\Vert\nabla F_n^l(\mathbf{w}_{n,\tau,j,t}^{l})\right\Vert^2\right\} + \frac{q_2}{N}(\tau+q_1\tau+\gamma) \mu^2 \frac{\sigma^2}{B}+\nonumber\\
		&\frac{(1+q_2)q_1}{N^2}\gamma \mu^2 \sum_{l}^{}     \sum_{n}^{} \sum_{j=0}^{\gamma-1}\mathbb{E}\left\{\left\Vert\nabla F_n^l(\mathbf{w}_{n,\tau,j,t}^{l})\right\Vert^2\right\}+\nonumber\\&\frac{(1+q_2)q_1}{N}\gamma \mu^2\frac{\sigma^2}{B} =\frac{\mu^2}{N}\left((\tau+\frac{q_1}{N})+\frac{q_2 q_1}{N}\right) \sum_{l}^{}  \sum_{i=0}^{\tau-1} \sum_{n \in {\cal C}^l}^{}\nonumber
						\end{align}
		\begin{align}
		& \mathbb{E}\left\{\left\Vert\nabla F_n^l(\mathbf{w}_{n,i,t}^l)\right\Vert^2\right\}+\frac{\mu^2}{N^2} \tau q_2 \sum_{l}^{}  N_l  \sum_{i=0}^{\tau-1}\sum_{n \in {\cal C}^l}^{} \mathbb{E}\Bigl\{ \nonumber\\
		&\left\Vert \nabla F_n^l(\mathbf{w}_{n,i,t}^l)\right\Vert^2\Bigr\}+\frac{\mu^2}{N}\gamma \left(1 +\frac{(1+q_2)q_1}{N} \right) \sum_{l}^{}  \sum_{n}^{} \sum_{j=0}^{\gamma-1}\nonumber\\
		&\mathbb{E}\left\{\left\Vert\nabla F_n^l(\mathbf{w}_{n,\tau,j,t}^{l})\right\Vert^2\right\}+\frac{q_2}{N^2}\gamma \mu^2 \sum_{l}^{}   N_l\sum_{n}^{} \sum_{j=0}^{\gamma-1}\nonumber\\&\mathbb{E}\left\{\left\Vert\nabla F_n^l(\mathbf{w}_{n,\tau,j,t}^{l})\right\Vert^2\right\}+\frac{\mu^2}{N} \frac{\sigma^2}{B}(\tau+\gamma)(1+q_2)\left(1+ {q_1}\right).
	\end{align}
	Finally, replacing \eqref{1RHS1} and \eqref{new_prod} in \eqref{curlexp}, and replacing the result with \eqref{normtwofinal} in \eqref{lipexp}, we have 
	\begin{align}
		\label{lipexp22}
		&\mathbb{E}\left\{F({\mathbf{w}}_{t+1})-F({\mathbf{w}}_{t})\right\} \leq -\frac{\mu \tau}{2} \mathbb{E}\left\{\|\nabla F( {\mathbf{w}}_{t})\|^2\right\}-\frac{\mu}{2N} \sum_{l}^{}\nonumber\\
		&\sum_{i=0}^{\tau-1}\sum_{n}^{}\mathbb{E}\left\{\|\nabla F_n^l(\mathbf{w}_{n,i,t}^l)\|^2\right\}+\frac{L^2\mu^3}{2N}\sum_{l} \sum_{i=0}^{\tau-1} \left({i+\frac{q_1}{N_l}}\right)\nonumber\\
		&\sum_{j=0}^{i-1} \sum_{n} \mathbb{E}\left\{\|\nabla F_n^l(\mathbf{w}_{n,j,t}^l)\|^2\right\} + \frac{L^2 \mu^3}{2N}C(1+q_1)\frac{\sigma^2}{B} \frac{\tau(\tau-1)}{2}\nonumber\\&+\frac{\mu \tau}{2}G^2-\frac{\mu\gamma}{2}  \mathbb{E}\left\{\|\nabla F( {\mathbf{w}}_{t})\|^2\right\}  -\mu\frac{1}{2N} \sum_{l}^{}\sum_{n}^{}\sum_{j=0}^{\gamma-1}\nonumber\\
		& \mathbb{E}\left\{\|\nabla F(\mathbf{w}_{n,\tau,j,t}^l)\|^2\right\} +\frac{L^2\mu^3}{2N} \tau \gamma \sum_{l}^{}   \sum_{i=0}^{\tau-1} \sum_{n \in {\cal C}^l}^{} \nonumber\\
		&\mathbb{E}\left\{\left\Vert\nabla F_n^l(\mathbf{w}_{n,i,t}^l)\right\Vert^2\right\} +\frac{L^2\mu^3}{2N} \sum_{l}^{}\sum_{n}^{}\sum_{j=0}^{\gamma-1}  j\sum_{p=0}^{j-1}\nonumber\\& \mathbb{E}\left\{\left\Vert\nabla F_n^l(\mathbf{w}_{n,\tau,p,t}^{l})\right\Vert^2\right\}+ \frac{L^2\mu^3}{2N} q_1\gamma \sum_{l}^{} \frac{1}{N_l}\sum_{i=0}^{\tau-1} \sum_{n \in {\cal C}^l}^{} \nonumber\\
		&\mathbb{E}\left\{\left\Vert \nabla F_n^l(\mathbf{w}_{n,i,t}^l)\right\Vert^2\right\}+\frac{L^2\mu^3}{2N} \tau \gamma C\frac{\sigma^2}{B}(1+q_1)+\frac{L^2\mu^3}{2} \frac{\sigma^2}{B}\nonumber\\
		& \frac{\gamma(\gamma-1)}{2}+\frac{L\mu^2}{2N}\left((\tau+\frac{q_1}{N})+\frac{q_2 q_1}{N}\right) \sum_{l}^{}  \sum_{i=0}^{\tau-1} \sum_{n \in {\cal C}^l}^{} \nonumber\\&\mathbb{E}\left\{\left\Vert\nabla F_n^l(\mathbf{w}_{n,i,t}^l)\right\Vert^2\right\}+\frac{L\mu^2}{2N^2} \tau q_2  \sum_{l}^{}  N_l  \sum_{i=0}^{\tau-1}\sum_{n \in {\cal C}^l}^{}  \nonumber\\
		&\mathbb{E}\left\{\left\Vert \nabla F_n^l(\mathbf{w}_{n,i,t}^l)\right\Vert^2\right\}+\frac{L\mu^2}{2N}\gamma \left(1 +\frac{(1+q_2)q_1}{N} \right) \sum_{l}^{}  \sum_{n}^{}\nonumber\\& \sum_{j=0}^{\gamma-1}\mathbb{E}\left\{\left\Vert\nabla F_n^l(\mathbf{w}_{n,\tau,j,t}^{l})\right\Vert^2\right\}+\frac{Lq_2}{2N^2}\gamma \mu^2 \sum_{l}^{}   N_l\sum_{n}^{} \sum_{j=0}^{\gamma-1}\nonumber\\
		&\mathbb{E}\left\{\left\Vert\nabla F_n^l(\mathbf{w}_{n,\tau,j,t}^{l})\right\Vert^2\right\}+\frac{L\mu^2}{2N} \frac{\sigma^2}{B}(\tau+\gamma)(1+q_2)\left(1+ {q_1}\right)\nonumber\\
		&+\frac{\mu \gamma}{2}G^2=-\frac{\mu (\tau+\gamma)}{2} \mathbb{E}\left\{\|\nabla F( {\mathbf{w}}_{t})\|^2\right\}-\frac{\mu}{2N}\biggl(1-{L^2\mu^2} \nonumber\\
		&\tau \gamma  -{L\mu}\left((\tau+\frac{q_1}{N})+\frac{q_2 q_1}{N}\right)\biggr) \sum_{l}^{}\sum_{i=0}^{\tau-1}\sum_{n}^{}\nonumber\\
		&\mathbb{E}\left\{\|\nabla F_n^l(\mathbf{w}_{n,i,t}^l)\|^2\right\}+\frac{L^2\mu^3}{2N}\sum_{l} \sum_{i=0}^{\tau-1} \left({i+\frac{q_1}{N_l}}\right)\sum_{j=0}^{i-1} \sum_{n}\nonumber
				\end{align}
		\begin{align}
		& \mathbb{E}\left\{\|\nabla F_n^l(\mathbf{w}_{n,j,t}^l)\|^2\right\}+ \frac{L^2\mu^3}{2N} q_1\gamma \sum_{l}^{} \frac{1}{N_l}\sum_{i=0}^{\tau-1} \sum_{n \in {\cal C}^l}^{} \nonumber\\
		&\mathbb{E}\left\{\left\Vert \nabla F_n^l(\mathbf{w}_{n,i,t}^l)\right\Vert^2\right\}+\frac{L\mu^2}{2N^2} \tau q_2 \sum_{l}^{}  N_l  \sum_{i=0}^{\tau-1}\sum_{n \in {\cal C}^l}^{}\mathbb{E}\bigl\{ \nonumber\\
		& \left\Vert \nabla F_n^l(\mathbf{w}_{n,i,t}^l)\right\Vert^2\bigr\}-\mu\frac{1}{2N}\left(1-{L\mu}\gamma \left(1 +\frac{(1+q_2)q_1}{N} \right)\right)\nonumber\\
		& \sum_{l}^{}\sum_{n}^{}\sum_{j=0}^{\gamma-1}\mathbb{E}\left\{\|\nabla F_n^l(\mathbf{w}_{n,\tau,j,t}^l)\|^2\right\} +\frac{Lq_2}{2N^2}\gamma \mu^2 \sum_{l}^{}   N_l\nonumber\\
		&\sum_{n}^{} \sum_{j=0}^{\gamma-1}\mathbb{E}\left\{\left\Vert\nabla F_n^l(\mathbf{w}_{n,\tau,j,t}^{l})\right\Vert^2\right\}+\frac{L^2\mu^3}{2N} \sum_{l}^{}\sum_{n}^{}\sum_{j=0}^{\gamma-1}  j\sum_{p=0}^{j-1} \nonumber\\&\mathbb{E}\left\{\left\Vert\nabla F_n^l(\mathbf{w}_{n,\tau,p,t}^{l})\right\Vert^2\right\}+ \frac{L\mu^2}{2}\frac{\sigma^2}{B}\biggl(\frac{L \mu}{N}C(1+q_1) \frac{\tau(\tau-1)}{2}\nonumber\\&+\frac{L\mu}{N} \tau \gamma C(1+q_1)+{L\mu}  \frac{\gamma(\gamma-1)}{2}+\nonumber\\&\frac{1}{N} (\tau+\gamma)(1+q_2)\left(1+ {q_1}\right)\biggr)+ \frac{\mu (\tau+\gamma)}{2}G^2.
	\end{align}
	Then, using the bounds
	\begin{align}
		&\sum_{i=0}^{\tau-1} i \sum_{j=0}^{i-1} \sum_{n}^{} \mathbb{E}\left\{\|\nabla F_n^l(\mathbf{w}_{n,j,t}^l)\|^2\right\} \leq \sum_{i=0}^{\tau-1} i \times \sum_{i=0}^{\tau-1} \sum_{n}^{} \nonumber\\
		&\mathbb{E}\bigl\{\|\nabla F_n^l(\mathbf{w}_{n,i,t}^l)\|^2\bigr\} = \frac{\tau(\tau-1)}{2}\times\nonumber\\
		&\sum_{i=0}^{\tau-1} \sum_{n}^{} \mathbb{E}\left\{\|\nabla F_n^l(\mathbf{w}_{n,i,t}^l)\|^2\right\},
	\end{align}
	\begin{align}
		&\sum_{j=0}^{\gamma-1}  j\sum_{p=0}^{j-1} \mathbb{E}\left\{\left\Vert\nabla F_n^l(\mathbf{w}_{n,\tau,p,t}^{l})\right\Vert^2\right\} \leq \frac{\gamma (\gamma-1)}{2} \sum_{j=0}^{\gamma-1}  \sum_{p=0}^{j-1} \nonumber\\
		&\mathbb{E}\left\{\left\Vert\nabla F_n^l(\mathbf{w}_{n,\tau,p,t}^{l})\right\Vert^2\right\},
	\end{align}
	\begin{align}
		&\sum_{l} \sum_{i=0}^{\tau-1} {\frac{1}{N_l}}\sum_{j=0}^{i-1} \sum_{n} \mathbb{E}\left\{\|\nabla F_n^l(\mathbf{w}_{n,j,t}^l)\|^2\right\} \leq \max_l{\frac{1}{N_l}}\times\nonumber\\&\sum_{l} \sum_{i=0}^{\tau-1} \sum_{j=0}^{i-1} \sum_{n} \mathbb{E}\left\{\|\nabla F_n^l(\mathbf{w}_{n,j,t}^l)\|^2\right\},
	\end{align}
	and
	\begin{align}
		&\sum_{l}^{}  N_l  \sum_{i=0}^{\tau-1}\sum_{n}^{}  \mathbb{E}\left\{\left\Vert \nabla F_n^l(\mathbf{w}_{n,i,t}^l)\right\Vert^2\right\} \leq \max_l N_l \times \nonumber\\&\sum_{l}^{} \sum_{i=0}^{\tau-1}\sum_{n}^{}  \mathbb{E}\left\{\left\Vert \nabla F_n^l(\mathbf{w}_{n,i,t}^l)\right\Vert^2\right\},
	\end{align}
	the following bound on \eqref{lipexp22} is obtaind.
	\begin{align}
		\label{lipexp2}
		&\mathbb{E}\left\{F({\mathbf{w}}_{t+1})-F({\mathbf{w}}_{t})\right\} \leq -\frac{\mu (\tau+\gamma)}{2} \mathbb{E}\left\{\|\nabla F( {\mathbf{w}}_{t})\|^2\right\}-\nonumber\\
		&\frac{\mu}{2N}\biggl(1-{L^2\mu^2} \tau \gamma  -{L\mu}\left((\tau+\frac{q_1}{N})+\frac{q_2 q_1}{N}\right)\biggr) \sum_{l}^{}\sum_{i=0}^{\tau-1}\sum_{n}^{}\nonumber\\
		&\mathbb{E}\left\{\|\nabla F_n^l(\mathbf{w}_{n,i,t}^l)\|^2\right\}+\frac{L^2\mu^3}{2N}\frac{\tau(\tau-1)}{2}\sum_{l}  \sum_{j=0}^{\tau-1} \sum_{n} \nonumber
						\end{align}
		\begin{align}
		&\mathbb{E}\left\{\|\nabla F_n^l(\mathbf{w}_{n,j,t}^l)\|^2\right\}+\frac{L^2\mu^3}{2N}q_1 \tau\max_{l}  {\frac{1}{N_l}} \sum_{l}\sum_{j=0}^{\tau-1} \sum_{n} \nonumber\\&\mathbb{E}\left\{\|\nabla F_n^l(\mathbf{w}_{n,j,t}^l)\|^2\right\}+ \frac{L^2\mu^3}{2N} q_1\gamma \max_{l}^{} \frac{1}{N_l}\sum_{l}^{}\sum_{i=0}^{\tau-1} \sum_{n \in {\cal C}^l}^{} \nonumber\\
		&\mathbb{E}\left\{\left\Vert \nabla F_n^l(\mathbf{w}_{n,i,t}^l)\right\Vert^2\right\}+\frac{L\mu^2}{2N^2} \tau q_2 \max_l N_l \sum_{l}\sum_{i=0}^{\tau-1}\sum_{n \in {\cal C}^l}^{}  \nonumber\\
		&\mathbb{E}\left\{\left\Vert \nabla F_n^l(\mathbf{w}_{n,i,t}^l)\right\Vert^2\right\}-\frac{\mu}{2N}\biggl(1-{L\mu}\gamma \left(1 +\frac{(1+q_2)q_1}{N} \right)\nonumber\\
		&\biggr) \sum_{l}^{}\sum_{n}^{}\sum_{j=0}^{\gamma-1} \mathbb{E}\left\{\|\nabla F_n^l(\mathbf{w}_{n,\tau,j,t}^l)\|^2\right\}+\frac{Lq_2}{2N^2}\gamma \mu^2 \max_l N_l  \nonumber\\
		& \sum_{l}\sum_{n}^{} \sum_{j=0}^{\gamma-1}\mathbb{E}\left\{\left\Vert\nabla F_n^l(\mathbf{w}_{n,\tau,j,t}^{l})\right\Vert^2\right\}+\frac{L^2\mu^3}{2N}\frac{\gamma(\gamma-1)}{2} \nonumber\\
		& \sum_{l}^{}\sum_{n}^{}\sum_{p=0}^{\gamma-1} \mathbb{E}\left\{\left\Vert\nabla F_n^l(\mathbf{w}_{n,\tau,p,t}^{l})\right\Vert^2\right\}+ \frac{L\mu^2}{2}\frac{\sigma^2}{B}\biggl(\frac{L \mu}{N}C\nonumber\\
		&(1+q_1) \frac{\tau(\tau-1)}{2}+\frac{L\mu}{N} \tau \gamma C(1+q_1)+{L\mu}  \frac{\gamma(\gamma-1)}{2}+\frac{1}{N}\nonumber\\
		& (\tau+\gamma)(1+q_2)\left(1+ {q_1}\right)\biggr)=-\frac{\mu (\tau+\gamma)}{2} \mathbb{E}\left\{\|\nabla F( {\mathbf{w}}_{t})\|^2\right\}\nonumber\\&-\frac{\mu}{2N} \biggl(1-{L^2\mu^2} (\tau \gamma+\frac{\tau(\tau-1)}{2}+q_1(\tau+\gamma)\max_{l}\frac{1}{N_l})  \nonumber\\
		&-{L\mu}\left((\tau+\frac{q_1}{N})+\frac{q_2 q_1}{N}+ \frac{\tau q_2\max_l N_l}{N} \right)\biggr) \sum_{l}^{}\sum_{i=0}^{\tau-1}\sum_{n}^{}\nonumber\\
		&\mathbb{E}\left\{\|\nabla F_n^l(\mathbf{w}_{n,i,t}^l)\|^2\right\}-\frac{\mu}{2N}\biggl(1-{L\mu}\gamma \biggl(1 +\frac{(1+q_2)q_1}{N}+\nonumber\\
		&\frac{q_2 \max_l N_l}{N} \biggr)-{L^2\mu^2}\frac{\gamma(\gamma-1)}{2}\biggr) \sum_{l}^{}\sum_{n}^{}\sum_{j=0}^{\gamma-1}\nonumber\\
		& \mathbb{E}\left\{\|\nabla F_n^l(\mathbf{w}_{n,\tau,j,t}^l)\|^2\right\} + \frac{L\mu^2}{2}\frac{\sigma^2}{B}\biggl(\frac{L \mu}{N}C(1+q_1) \frac{\tau(\tau-1)}{2}\nonumber\\&+\frac{L\mu_t}{N} \tau \gamma C(1+q_1)+{L\mu}  \frac{\gamma(\gamma-1)}{2}+\frac{1}{N} (\tau+\gamma)(1+q_2)\nonumber\\&\left(1+ {q_1}\right)\biggr) + \frac{\mu (\tau+\gamma)}{2}G^2.
	\end{align}
	Given the conditions
	\begin{align}
		&1-{L^2\mu^2} (\tau \gamma+\frac{\tau(\tau-1)}{2}+q_1(\tau+\gamma)\max_{l}\frac{1}{N_l})  -{L\mu}\nonumber\\&\left((\tau+\frac{q_1}{N})+\frac{q_2 q_1}{N}+ \frac{\tau q_2\max_l N_l}{N} \right) \geq 0, 
	\end{align}
	and
	\begin{align}
		&1-{L\mu}\gamma \left(1 +\frac{(1+q_2)q_1}{N}+\frac{q_2\max_l N_l}{N} \right)-\nonumber\\&{L^2\mu^2}\frac{\gamma(\gamma-1)}{2} \geq 0,
	\end{align}
	the second and third terms in RHS of \eqref{lipexp2} are negative, and
	after applying Assumption 4, we can write for any $t \in \left\{0,\cdots,T-1\right\}$ 
	\begin{align}
		\label{proof_final_step}
		&\mathbb{E}\left\{F({\mathbf{w}}_{t+1})\right\}-F^* \leq (1-\mu \delta (\tau+\gamma))(\mathbb{E}\left\{F({\mathbf{w}}_{t})\right\}-F^*)\nonumber\\&+ \frac{L\mu^2}{2}\frac{\sigma^2}{B}\biggl(\frac{L \mu}{N}C(1+q_1) \frac{\tau(\tau-1)}{2}+\frac{L\mu}{N} \tau \gamma C(1+q_1)+\nonumber\\&{L\mu}  \frac{\gamma(\gamma-1)}{2}+\frac{1}{N} (\tau+\gamma)(1+q_2)\left(1+ {q_1}\right)\biggr) + \frac{\mu (\tau+\gamma)}{2}G^2.
	\end{align}
	This bound links the steps $t+1$ and $t$. To determine the bound
	specified in Theorem 1, we can substitute
	$\mathbb{E}\left\{F({\mathbf{w}}_{t})\right\}-F^*$ on the RHS with the equivalent one-step bound for the
	steps $t$ and $t-1$. By consistently applying this procedure over
	the interval $\left\{t-1,\ldots,0\right\}$, the proof is complete.

\end{document}